\documentclass[twoside]{article}

\usepackage[accepted]{aistats2025}
\usepackage[round]{natbib}

\usepackage{amsmath, amsthm }
\usepackage[utf8]{inputenc} 
\usepackage[T1]{fontenc}    
\usepackage[backref=page, colorlinks = True, linkcolor = mydarkblue, citecolor = mydarkblue]{hyperref}
\usepackage{cleveref}
\usepackage{url}            
\usepackage{booktabs}       
\usepackage{amsfonts}       
\usepackage{nicefrac}       
\usepackage{microtype}      
\usepackage{xcolor}         
\usepackage{multirow}
\usepackage{wrapfig}

\usepackage[makeroom]{cancel}
\usepackage{graphicx}
\usepackage{svg}

\usepackage{caption}
\usepackage{subfigure}
\usepackage{algpseudocode}
\usepackage{algorithm}

\usepackage{multicol}

\newtheorem{proposition}{Proposition}

\crefname{assumption}{}{}

\crefname{oldassumption}{}{}

\def\X{\mathcal{X}}

\definecolor{mydarkblue}{rgb}{0,0.08,0.45}

\newcommand{\x}{x}
\newcommand{\xo}{x_\mathrm{o}}
\newcommand{\btheta}{\theta}

\newcommand{\bphi}{\phi}

\newcommand{\cond}{\, | \, }
\newcommand{\proposal}{\tilde{\pi}_g}

\newenvironment{talign*}
 {\csname align*\endcsname}
 {\endalign}

\begin{document}

%

%

\twocolumn[

\aistatstitle{Cost-aware simulation-based inference}

\aistatsauthor{ Ayush Bharti \And Daolang Huang \And  Samuel Kaski \And Fran\c{c}ois-Xavier Briol }

\aistatsaddress{ Aalto University \And  Aalto University \And Aalto University\\ University of Manchester \And University College London } ]

\begin{abstract}
  Simulation-based inference (SBI) is rapidly becoming the preferred framework for estimating parameters of intractable models in science and engineering. A significant challenge in this context is the large computational cost of simulating data from complex models, and the fact that this cost often depends on parameter values.
  We therefore propose \textit{cost-aware SBI methods} which can significantly reduce the cost of existing sampling-based SBI methods, such as neural SBI and approximate Bayesian computation. 
  This is achieved through a combination of rejection and self-normalised importance sampling, which reduces the number of expensive simulations needed. Our approach is studied extensively on models from epidemiology to telecommunications engineering, where we obtain significant reductions in the overall cost of inference.
\end{abstract}

\section{INTRODUCTION}
\label{sec:introduction}

Many scientific disciplines use computational and mechanistic models to study complex physical, biological, or sociological phenomena. Such models, or \emph{simulators}, are often poorly suited for standard Bayesian inference methods due to their intractable likelihood functions. Simulation-based inference (SBI; \cite{Cranmer2020}) addresses this issue by approximating posteriors using repeated simulations from the model instead of evaluations of the likelihood. These simulations are typically used to measure similarity with the observed data \citep{Lintusaari2016, Sisson2018, Briol2019MMD}, or for training a conditional density estimator \citep{Lueckmann2021, zammitmangion2024neural}, and a large number of these is therefore required. A significant bottleneck limiting the use of SBI in many real-world cases is therefore the cost of simulators. Even if a model takes only a minute for a single simulation, state-of-the-art SBI often requires tens of thousands of simulations which can take several days to run. Needless to say, costlier simulators taking hours or days per simulation, such as most tsunami models \citep{Behrens2015}, wind farm models \citep{Kirby2023} or nuclear fusion simulators \citep{DREAM}, are currently out of the reach of SBI.

  \begin{figure}
      \centering
    \includegraphics[width=0.75\columnwidth]{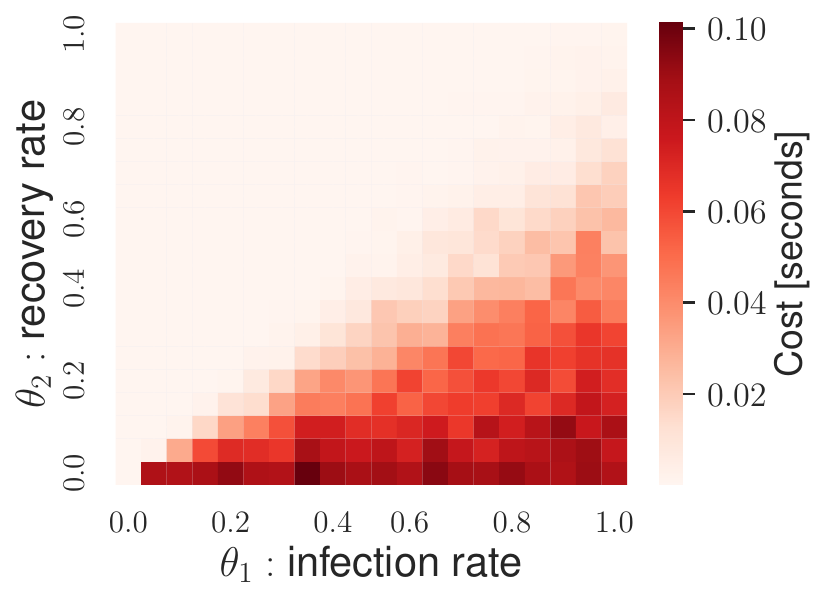}
      \caption{Estimated cost of the temporal susceptible-infected-recovered  model considered in \Cref{sec:experiments_epidemiology}.}
      \label{fig:intro}
      \vspace{-3mm}
  \end{figure}

Surprisingly, existing SBI methods tend to ignore the fact that the cost of a simulator will often depend on the parameter value for which we are simulating, and are therefore unable to take advantage of this property. For example, many models of disease spread in epidemiology have varying cost \citep{Kypraios2017,McKinley2018}; as the recovery rate decreases and the infection rate increases, the disease spreads faster, and the computational cost of a simulation increases. This is illustrated in \Cref{fig:intro} for a model we will study in \Cref{sec:experiments_epidemiology}, in which case some parametrisations are up to $10$ times more expensive than cheaper parametrisations. Varying cost is also a feature in many other real-world applications, including aircraft vehicle design simulators \citep{Cobb2023} where simulating aircraft with large number of propellers and wings is slower, and graph-based radio propagation models where simulating graphs with more nodes takes longer \citep{Bharti2022}. 

In this paper, we tackle this issue by proposing the first family of \textit{cost-aware} alternatives to popular SBI methods such as neural posterior estimation (NPE) \citep{Papamakarios2016, Radev2022}, neural likelihood estimation (NLE) \citep{Papamakarios2019} and approximate Bayesian computation (ABC) \citep{Sisson2018}. Cost-aware SBI uses self-normalised importance sampling with an importance distribution constructed to encourage sampling from the cheaper parameterisations of the model. This leads to SBI methods capable of significant computational savings without compromising on accuracy. We demonstrate these advantages on real-world simulators from the fields of epidemiology and radio propagation, leading to significant reductions in the cost of inference in both cases. Interestingly, cost-aware SBI complements, rather than replaces, many of the existing approaches for sample-efficiency in the literature.

\section{BACKGROUND} \label{sec:background}
Consider a parametric model (i.e. a family of Borel probability distributions) on some space $\mathcal{X} \subseteq \mathbb{R}^d$ with parameters $\theta \in \Theta \subseteq \mathbb{R}^p$. We will assume this is a simulator-based model, meaning that the density $p(\cdot \cond \btheta)$ (and therefore the likelihood) associated with this model cannot be evaluated point-wise, but it is possible to simulate independent realisations $\x \sim p(\x \cond \btheta)$\footnote{We abuse notation by writing $\x \sim p(\x \cond \theta)$ for independent sampling from the distribution with density $p(\cdot \cond \btheta)$.} for a fixed $\btheta \in \Theta$ (albeit at a potentially significant computational cost). Given $m_o$ observations $x_{o} = (x_{o1},\ldots,x_{om_o}) \in \mathcal{X}^{m_o}$ and a prior $p(\btheta)$, we are interested in the posterior
\begin{equation*}
    p(\btheta \cond \xo) = \frac{p(\xo \cond \btheta) p(\btheta)}{\int_\Theta p(\xo \cond \btheta) p(\btheta) d \theta} \propto p(\xo \cond \btheta) p(\btheta),
\end{equation*}
where $\propto$ denotes proportionality up to a multiplicative constant. As the likelihood is intractable, SBI methods rely on simulations from the joint $(x, \theta) \sim p(x, \theta) = p(x|\theta)p(\theta)$ to approximate the posterior distribution.
This is typically achieved by first sampling from the prior $\theta \sim p(\theta)$ and then simulating $m$ data points from the model $(x_1, \dots, x_m) \sim p(\x \cond \btheta)$. Suppose the expected cost of simulating a single realisation from the model with parameter $\theta$ is given by $c(\theta)$, where $c: \Theta \rightarrow (0,\infty)$ is called the \emph{cost function}. Then, the total expected cost of simulating $m$ realisations from $n$ parameter values $\theta_1, \ldots, \theta_n$ becomes $\sum_{i=1}^n c(\theta_i)m$. In this paper, the cost function will usually refer to computational cost (measured in units of time), but it could conceptually represent other forms of cost such as financial cost or memory cost. 
We now introduce some popular SBI methods, emphasising the computationally costly operations in each method. 

\paragraph{Neural posterior estimation (NPE).} NPE \citep{Radev2022} is a neural SBI method that uses a conditional density estimator, such as a normalising flow \citep{papamakarios2021normalizing}, to learn an approximate mapping from data to the posterior: $x \mapsto p(\btheta \cond \x)$. Let $q_\phi(\theta \cond x)$ denote such a conditional normalising flow parameterised by the vector $\phi$. 
The parameter $\phi$ can be estimated by minimising the empirical loss
\begin{align} \label{eq:npe_loss}
    \ell_{\text{NPE}}(\bphi) &=  -\frac{1}{m n} \sum_{j=1}^m \sum_{i=1}^n \log q_{\phi} (\btheta_i \cond x_{ij})\nonumber \\
    &\approx -\mathbb{E}_{\theta \sim p(\theta)}[\mathbb{E}_{x\sim p(x \cond \theta)}[\log q_{\phi} (\theta \cond x)]] ,
\end{align}
 using the dataset $\{(\btheta_i, \{\x_{ij}\}_{j=1}^m)\}_{i=1}^n$ generated from the joint.
After training, inference is amortised; given the estimated flow parameters $\hat{\phi}$ and a new observed dataset $\xo$, an approximation of the posterior is given by simply evaluating $p_{\text{NPE}}(\btheta \cond \xo) := q_{\hat{\phi}}(\btheta \cond \xo)$. However, training $q_\phi$ requires many realisations from the simulator in order to accurately approximate the expected loss, which can be a significant bottleneck if the simulator is computationally expensive.

\paragraph{Neural likelihood estimation (NLE).} 
In NLE \citep{Papamakarios2019}, a conditional density estimator  is used to learn a mapping from the parameters to the likelihood: $\btheta \mapsto p(\x \cond \btheta)$. Given simulations from the joint, the density estimator $q_\phi(x \cond \theta)$ is trained by minimising the following empirical loss
\begin{align}\label{eq:nle_loss}
    \ell_{\text{NLE}}(\phi) &=  -\frac{1}{m n} \sum_{j=1}^m \sum_{i=1}^n \log q_\phi(x_{ij} \cond \theta_i) \nonumber \\
    &\approx -\mathbb{E}_{\theta \sim p(\theta)}[\mathbb{E}_{x\sim p(x \cond \theta)}[\log q_\phi(x \cond \theta)]].
\end{align}
The trained model is used to obtain an approximate posterior $p_{\mathrm{NLE}}(\theta \cond \xo) \propto q_{\hat{\phi}}(x_o \cond \theta)p(\theta)$ which can typically be sampled from using Markov chain Monte Carlo (MCMC). 
Once again, the costly step is simulating the realisations used to estimate the expected loss.

\paragraph{Approximate Bayesian computation (ABC).} ABC \citep{Lintusaari2016, Sisson2018} is an SBI framework that relies on a distance $\varrho: \X^{m_o} \times \X^m \rightarrow [0,\infty)$ to create an approximate posterior
\begin{align*}
    p_{\text{ABC}}(\btheta \cond \xo) \propto \mathbb{E}_{\x\sim p(\x \cond \btheta)}[1_{\{\varrho(\xo,\x) \leq \epsilon\}}]p(\btheta).
\end{align*}
The classic accept-reject ABC method \citep{Pritchard1999} to approximate $p_{\text{ABC}}$ involves repeatedly sampling $\btheta \sim p(\btheta)$ from the prior, simulating data $\x \sim p(\x \cond\btheta)$ using the simulator, and then accepting $\btheta$ if the distance $\varrho( \xo,\x)$ falls below a tolerance threshold $\epsilon > 0$. 
Even the more advanced ABC methods based on sequential Monte Carlo  \citep{Beaumont2009, DelMoral2011} and regression adjustment \citep{Beaumont2002} rely on this two-step process. 
To ensure ABC provides an accurate approximation of $p(\theta|\xo)$, we must take $\epsilon$ to be small \citep{Frazier2018}, but this comes at the cost of a high rejection rate which increases the number of simulator calls. ABC can therefore be very computationally demanding when dealing with expensive simulators.

\paragraph{Related work.}
To tackle the challenge of expensive simulations, a number of \textit{sample-efficient} SBI methods have been proposed. 
A full review is beyond the scope of this paper, but these approaches can be broadly categorised as follows: \textit{(i)} methods that focus on improved sampling of the posterior, hence reducing the total number of parameters for which simulations are needed e.g. through sequential sampling \citep{Sisson2007, Beaumont2009, DelMoral2011,Greenberg2019, Papamakarios2019, Hermans2020} or Gaussian process surrogates \citep{meeds2014gps, wilkinson2014accelerating, Gutmann2016, Jarvenpaa2019}; \textit{(ii)} methods that improve upon independent simulations of data given a fixed parameter value, e.g. using quasi-Monte Carlo or Bayesian quadrature \citep{Niu2021,Bharti2023}, hence reducing the number of samples needed per parameter value; and \textit{(iii)} methods using side-information to avoid unnecessary simulations, e.g. multi-fidelity methods \citep{Prescott2020, Prescott2021, warne2022multifidelity, prescott2024efficient}, methods based on partial simulations \citep{Prangle2016}, and expert-in-the-loop approaches \citep{Bharti2022}. Our proposed cost-aware sampling method falls in the third category as it uses knowledge of the cost of simulations to guide sampling in the parameter space. Importantly, it is complementary to all of the approaches described above, including methods that parallelise computations \citep{kulkarni2023improving}.
Note that importance sampling has previously been used in the context of SBI by \citet{Dax2023, Prangle2023}, but their focus was solely on variance reduction rather than on reducing the computational cost without compromising on the variance.

\section{METHOD}
\label{sec:method}

We now describe cost-aware importance sampling in \Cref{sec:ca_sampling}, discuss practical considerations in Sections \ref{sec:cost_efficiency_tradeoff} and \ref{sec:choosing_g}, then present cost-aware SBI in \Cref{sec:ca_sbi}. 

\subsection{Cost-aware importance sampling}
\label{sec:ca_sampling}

Let $\pi$ be a target density on $\Theta \subseteq \mathbb{R}^p$. In the context of SBI, $\pi$ will either be the prior or the approximate posterior, but for now it can be considered  arbitrary. We assume that sampling independent realisations $\theta \sim \pi$ is cheap, however, each sampled $\theta$ will be used for a downstream task with expected cost $c(\theta)$. In SBI, $c(\theta)$ will be the expected cost of simulating from the simulator. In scenarios where the cost varies across $\Theta$, the standard approach of sampling $n$ times independently from $\pi$ (in a cost-agnostic manner) can lead to a large downstream cost. We therefore aim to reduce the downstream cost by using $c(\theta)$ to guide our sampling towards cheaper regions of  $\Theta$. Instead of sampling independently from the target $\pi(\theta)$, we propose to use a \textit{cost-aware proposal} distribution with density
\begin{equation*}
    \proposal(\theta) \propto \frac{\pi(\theta)}{g(c(\theta))},
\end{equation*}
and re-weight samples so as to approximate $\pi$. Here, $g:(0,\infty) \rightarrow (0,\infty)$ is a non-decreasing function that governs the degree to which we penalise for cost, and the composition $g \circ c$ must be strictly positive. The assumption that $g$ is non-decreasing is needed to ensure that larger costs are penalised at least as much as smaller costs, and the assumption that $g \circ c$ is strictly positive can trivially be satisfied for any cost function which is strictly positive. The function $g$ is a user-specified choice, and we discuss its selection in \Cref{sec:choosing_g}. Interestingly, when $g$ is a constant function, our approach does not penalise for cost and therefore reduces to i.i.d. sampling (and hence to standard SBI methods when used in that context). More generally, $\proposal$ can be interpreted as a form of exponential tilting \citep{siegmund1976importance} of $\pi$ with the function $\exp{(- \log g(c(\theta)))}$, but we are not aware of any focus on computational cost of a downstream task in this literature. The closest approach is that of Neyman allocation \citep{neyman1934two}, which solves the stratification problem when sampling cost varies; however, the notion of cost in this setting is that of sampling from $\pi$, rather than the cost of a downstream task.

Outside of a few specific choices of $\pi$, $c$, and $g$, we will typically not have access to closed-form expressions for the normalisation constant of the proposal $\proposal$ (see \Cref{tab:B_expression} in \Cref{sec:app_closedforms} for a few exceptions where we do).
However, this does not create an issue for sampling from $\proposal$, which we propose to do via rejection sampling \citep{neumann1951various} with proposal $\pi$. Samples from $\proposal$ are generated by first sampling candidates $\theta \sim \pi$, and then accepting those candidates with probability $A(\theta) = \proposal(\theta) / M \pi(\theta)$, where $M>0$ is a constant that satisfies the condition $\proposal(\theta) \leq M \pi(\theta)$. The following proposition formalises the conditions under which the cost-aware proposal $\proposal$ is a valid density, and gives an expression for $A(\theta)$. The proof is in \Cref{app:proof_rejection}.
\begin{proposition} \label{prop:rejection_sampling}
    Let $\pi$ be a density function on $\Theta$ and assume $g:(0,\infty) \rightarrow (0,\infty)$ is a non-decreasing function and $g \circ c$ is strictly positive; i.e. $g_{\mathrm{min}} := \inf_{\btheta \in \Theta} g(c(\theta))>0$. Then, $\proposal$ is also a density function.
    
    In addition, we can sample from $\proposal$ using rejection sampling with proposal $\pi$ and acceptance probability
    \begin{equation}
    \label{eq:acceptance_prob}
        A(\theta) = \frac{g_{\mathrm{min}} }{g(c(\btheta))}.
    \end{equation}
\end{proposition}
This rejection sampling algorithm is very widely applicable since there are no restrictions on $\pi$, and the conditions on $g$ and $g \circ c$ are minimal as previously discussed. In addition, the acceptance probability $A(\theta)$ does not require knowledge of the normalisation constant of $\proposal$. 
It only depends on $c$, which we assume known, and $g_{\text{min}}$, which is available since we can choose $g$. In particular, it does not rely on any costly down-stream task, making the approach computationally cheap. Alternative proposals, such as those used in adaptive rejection sampling \citep{Gilks1992, Gilks1995}, may provide an algorithm with slightly smaller rejection rate, but the tractability of the acceptance probability makes this approach particularly convenient. 

We note that when the minima of $g$ and $c$ are small, then $g_{\mathrm{min}}$ may also be small. This means that $A(\theta)$ will approach zero, leading to a higher rejection rate. The cost of sampling from $\proposal$ will increase in such scenarios, but this is not a significant issue as, for SBI, $\pi$ usually corresponds to a simple distribution such as a uniform or a Gaussian, which is extremely cheap to sample from. Furthermore, this issue can also be alleviated by choosing the penalty function to be of the form $g(c(\theta))=\max(1,h(c(\theta)))$ for some non-decreasing function $h$, which ensures that $ g_{\mathrm{min}}=1$, and hence $A(\theta)$ is not too close to zero. 

Now that we can sample from $\proposal$, we can replace samples from $\pi$ with those from $\proposal$ to minimise the total downstream cost. The fact that we samples from $\proposal$ instead of $\pi$ can be accounted for by weighting the realisations using self-normalised importance sampling \citep{trotter1956conditional}. The unnormalised weights in this case can be computed as
\begin{align}
    w(\theta) = \frac{\pi(\theta)}{\proposal(\theta)} = \frac{B \pi(\theta) g(c(\theta))}{\pi(\theta)} \propto g(c(\theta)),
\end{align}
where $B$ is the normalisation constant for $\proposal$.
Given realisations $\{\theta_i\}_{i=1}^n$ from $\proposal$, normalised weights can be obtained from the unnormalised weights as follows:
\begin{align}\label{eq:ca_weights}
    w_{\mathrm{Ca}}(\theta_i) = \frac{w(\theta_i)}{\sum_{j=1}^n w(\theta_j)} = \frac{g(c(\theta_i))}{\sum_{j=1}^n g(c(\theta_j))}.
\end{align}
The advantage of using self-normalisation is that the weights do not depend on the normalisation constants of either the proposal $\proposal$ or the target $\pi$.

Now that we have our cost-aware importance sampling scheme for $\pi$, we may be interested in computing the expected value of an arbitrary integrand $f:\Theta \rightarrow \mathbb{R}$ with respect to $\pi$, i.e., $\mu = \int_\Theta f(\theta) \pi(\theta) d\theta$. This can be achieved using the estimator $\hat{\mu}^{\text{Ca}}_n = \sum_{i=1}^n w_{\mathrm{Ca}}(\theta_i)f(\theta_i)$, which is consistent and has bounded weights and finite variance $\sigma^2_{\text{Ca}}$ under very mild conditions. In addition, we can relate the variance of this estimator (a key measure of its efficiency) to the variance $\sigma^2_{\text{MC}}$ of the Monte Carlo estimator $\hat{\mu}^\text{MC}_n = \frac{1}{n} \sum_{i=1}^n f(\theta_i)$; see \Cref{app:proof_variance} for the proof. 
\begin{proposition} \label{prop:finite_variance}
    Suppose $g$ is non-decreasing and $g_{\mathrm{min}}>0$. Then, $\hat{\mu}^{\text{Ca}}_n \rightarrow \mu$ with probability 1 as $n\rightarrow \infty$. 
    
    Furthermore, assume $g_{\mathrm{max}} =\sup_{\theta \in \Theta} g(c(\theta))  <\infty$ and $f$ is $\pi$-square-integrable ($\int_{\Theta} f(\theta)^2 \pi(\theta) d\theta < \infty$). Then 
    \begin{align*}
        \frac{g_{\text{min}}}{n g_{\text{max}}}  \leq  w_{\text{Ca}}(\theta_i) \leq \frac{g_{\text{max}}}{n g_{\text{min}}} \qquad \forall i \in \{1,\ldots,n\},
    \end{align*}
    and we have that
    \begin{align*}
    \frac{g_{\text{min}}}{g_{\text{max}}} \left(\sigma^2_{\text{MC}} - \frac{\mu^2}{n}\right) \leq \sigma^2_{\text{Ca}} \leq \frac{g_{\text{max}}}{g_{\text{min}}} \left(\sigma^2_{\text{MC}} - \frac{\mu^2}n \right).
    \end{align*}
\end{proposition}
 The assumptions are again extremely minimal. Assuming $f$ is $\pi$-square-integrable is standard and required for Monte Carlo to have finite variance. The assumption that $g_{\mathrm{max}} <\infty$ holds when $c$ is upper bounded, or can be enforced by choosing $g$ of the form $g(c(\theta)) = \min(h(c(\theta)), C)$, where $h$ is a strictly-positive non-decreasing function and $C$ is some large constant.

A direct corollary of this result is that the cost-aware estimator does not have infinite variance, which is a regular issue for importance sampling (see Chapter 9 of \cite{mcbook}). It also implies that cost-aware sampling is at worst $g_{\text{max}}/g_{\text{min}}$ less efficient than Monte Carlo, meaning we would like $g_{\text{min}}/g_{\text{max}}$ to be large. 

As a side note, we remark that we could be tempted to use the optimal self-normalised importance sampling proposal; i.e. the proposal which minimises the variance. However, this is  proportional to $|f(\theta) - \mu|\pi(\theta)$ \citep[Ch. 2]{hesterberg1988advances}, which corresponds to having $g(c(\theta)) \propto |f(\theta) - \mu|^{-1}$ and is sadly not a non-decreasing function. This choice is therefore not a valid proposal for our framework. 

Another alternative measure of efficiency is the \emph{effective sample size (ESS)} \citep[Section 9.3]{mcbook}:
\begin{align*}
    \text{ESS} = \frac{\left( \sum_{i=1}^n w(\btheta_i) \right)^2}{n \sum_{i=1}^n w(\btheta_i)^2} = \frac{\left( \sum_{i=1}^n g(c(\btheta_i)) \right)^2}{n\sum_{i=1}^n g(c(\btheta_i))^2}.
\end{align*}
This is an attractive measure since the variance is integrand-specific but the ESS is not.
Again, we can bound this quantity using our minimal assumptions; see \Cref{app:proof_ESS_bounds} for the proof.
\begin{proposition}\label{prop:bound_ESS}
    Suppose $g$ is non-decreasing and $0< g_{\mathrm{min}} \leq g_{\mathrm{max}}<\infty$. Then: 
    \begin{align*}
        \left(\frac{g_{\text{min}}}{g_{\text{max}}}\right)^2 \leq \text{ESS} \leq \left(\frac{g_{\text{max}}}{g_{\text{min}}}\right)^2.
    \end{align*}
\end{proposition}
Similarly to the variance result, efficiency (as measured by the ESS) can potentially be improved by taking the ratio $g_{\text{min}}/g_{\text{max}}$ as large as possible.

\subsection{Cost-versus-efficiency trade-off}\label{sec:cost_efficiency_tradeoff}

The penalty function $g$ can have a significant impact on cost-aware importance sampling since it leads to a cost-versus-efficiency trade-off. For efficiency, we have already seen that we should choose $g$ so as to maximise $g_{\text{min}}/g_{\text{max}}$. Of course, this is a simplistic view as it does not take cost into consideration.
We therefore also introduce the notion of \emph{computational gain (CG)}, which is the ratio of the expected cost of the downstream task using Monte Carlo to that of using cost-aware sampling:
\begin{equation*}
    \text{CG} = \frac{\int_\Theta c(\btheta) \pi(\btheta) d\btheta}{\int_\Theta c(\btheta) \proposal(\btheta) d\btheta}.  
\end{equation*}
The CG allows us to ascertain the reduction in cost we can expect to have given a particular choice of $g$; for instance, CG $=2$ implies a 50\% reduction in cost. The CG can be straightforwardly estimated using Monte Carlo samples from $\pi$ and $\proposal$ prior to running any expensive downstream tasks (and in some rare cases can even be obtained in closed-form; see \Cref{app:CG_derivation}). We now bound the CG; see \Cref{app:proof_CG_bounds} for the proof.
\begin{proposition}\label{prop:CG_bounds}
Suppose $g$ is non-decreasing and $0 < g_{\mathrm{min}} \leq g_{\mathrm{max}}<\infty$. Then: $1 \leq \text{CG} \leq \frac{g_{\text{max}}}{g_{\text{min}}}$.
\end{proposition}
Since we want to minimise the cost, or equivalently maximise the CG, we should maximise $g_\text{max}/g_\text{min}$, or equivalently minimise $g_{\text{min}}/g_{\text{max}}$. Unfortunately, this is in direct contradiction to our findings on maximising the ESS, which demonstrates a clear trade-off. Note that having a lower bound for the CG of $1$ guarantees that we will never increase our expected cost when using cost-aware importance sampling.

\subsection{Choosing the penalty function}\label{sec:choosing_g}

Given the cost-versus-efficiency trade-off above, the question of how to choose $g$ remains. Since we do not require running expensive simulations from the model to compute either the CG or the ESS, we propose to use these as a criterion. More precisely, we select $g$ such that the quantity CG$\times$ESS is above or close to $1$ (as would be the case for standard SBI methods). This ensures that the reduction in cost does not come at the expense of significant degradation in performance.

In cases where the expensive regions of $\Theta$ are important to sample from to obtain good efficiency (for instance, in SBI when the true posterior lies in the computationally costly region), choosing a single penalty function based on CG and ESS alone may lead to sub-optimal results. Therefore, we recommend an approach based on \emph{multiple importance sampling} (i.e. combining several importance sampling estimators), with components inspired by the defensive mixture approach of \citet{hesterberg1995weighted}.  We consider $J$ importance distributions, which include the target $\tilde{\pi}_1 = \pi$ and $J-1$ cost-aware proposals  $\tilde{\pi}_{j}(\theta) \propto \pi(\theta) / g_j(c(\theta))$ for $j=2,\ldots,J-1$. Given $\theta_{ij} \sim \tilde{\pi}_j$ for $i=1,\ldots,n_j$ and $j=1,\ldots,J$, an integral $\mu = \int_{\Theta} f(\theta) \pi(\theta) d\theta$ can be estimated using \emph{multiple cost-aware (mCa) importance sampling} as
\begin{align*}
\hat{\mu}^{\text{mCa}} = \frac{1}{J} \sum_{j=1}^J \sum_{i=1}^{n_j} w_{\text{Ca},j}(\theta_{ij}) f(\theta_{ij})
\end{align*}
where $w_{\text{Ca},j}$ corresponds to the weights in \Cref{eq:ca_weights} computed with penalisation function $g_j$. In practice, we found $n_1=\ldots=n_J$ and $J=4$ to work well and propose to use penalisation functions of the form $g_j(c(\theta))=c(\theta)^{k_j}$. A few of the components $k_j$ can be selected based on the CG$\times$ESS metric, while the others can be higher powers. By doing so, we get some samples from the expensive regions, whilst also getting the cost benefits from using larger values of $k_j$. 

Note that \citet{csonka2001cost} also considered cost for multiple importance sampling, but their notion of cost refers to the cost of sampling from an importance distribution rather than the cost of a downstream task. Their method is therefore not applicable to our setting.

\subsection{Cost-aware simulation-based inference}
\label{sec:ca_sbi}

\begin{algorithm}[t] 
\caption{Cost-aware SBI}
\label{alg:ca-sbi}
\begin{algorithmic}
\item \textbf{Input:} Number of samples $n$, target $p$, penalty function $g$, observed data $\xo$.
\State [Optional: Cost function $c$. If not available, replace $c$ by an estimate $\hat c$ in the algorithm below.]
\State Sample parameters $\btheta_1, \dots, \btheta_n$ from the cost-aware proposal $\tilde p_g$ through rejection sampling using $c$. 
\State Simulate from the model: $\x_i \sim p(\x \cond \btheta_i)$, $i=1,\dots, n$.
\State Perform SBI with real data $\xo$, simulated data $\{(\btheta_i, \{\x_{ij}\}_{j=1}^m)\}_{i=1}^n$ and weights $\{w_{\text{Ca}}(\theta_i)\}_{i=1}^n$.
\item \textbf{Output:} Weighted samples from the SBI posterior.
\end{algorithmic}
\end{algorithm}

\begin{figure}[t]
    \centering
    \includegraphics[trim={230 170 170 195}, clip,width=\columnwidth]{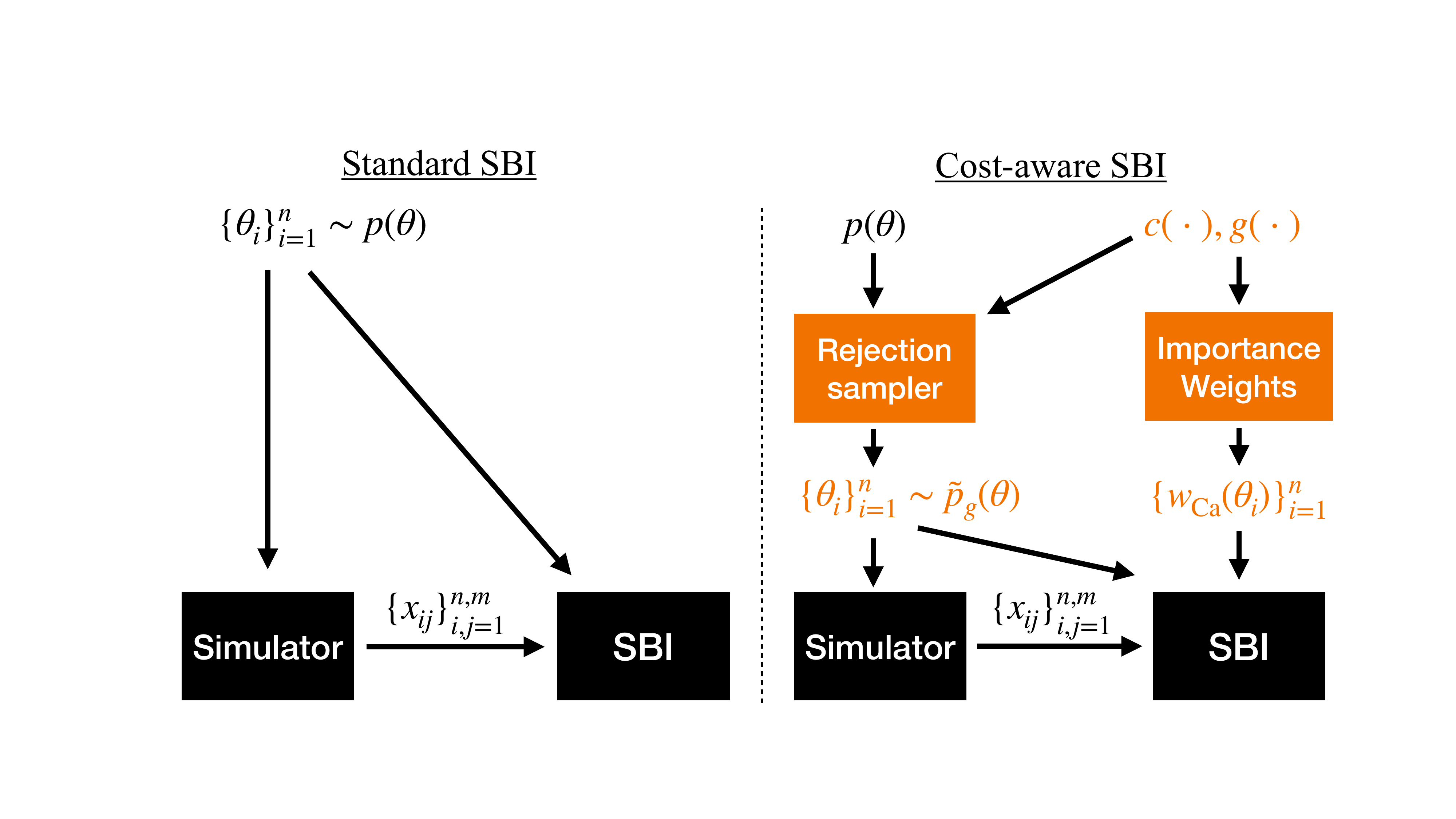}
    \vspace{-2mm}
    \caption{Flowchart of SBI and Ca-SBI. Ca-SBI utilises the cost function $c$ and a penalty function $g$ to (i) sample from the cost-aware proposal $\tilde p_g$, and (ii) compute the cost-aware weights. Step 1 reduces the overall cost from using the simulator, while step 2 guarantees we are sampling from the target SBI posterior.} 
    \label{fig:casbi_flowchart}
\end{figure}

We are now ready to apply our cost-aware sampling strategy to SBI. Here, $\pi$ typically corresponds to the prior or posterior, and $c(\theta)$ is the expected cost of a single simulation from the model with parameter value $\theta$, and will be measured using runtime on a given machine. Our approach encourages sampling parameter values corresponding to cheaper parametrisations of the simulator, and hence reduces the overall cost of SBI. It is summarised in \Cref{alg:ca-sbi} and \Cref{fig:casbi_flowchart}.

If $\theta \mapsto c(\btheta)$ is not known \textit{a priori}, we can estimate it by evaluating the simulator once for a few values of the parameters, say $(\btheta_1, \dots, \btheta_{\tilde{n}})$, $\tilde{n} \ll n$, and recording the computational time $\{y_i\}_{i=1}^{\tilde{n}}$. We can see these measurements as noisy evaluations of the expected cost: $y_i = c(\btheta_i) + \varepsilon_i$. By fitting a simple model such as a polynomial or a Gaussian process \citep{Rasmussen2006} using $\{(\btheta_i, y_i)\}_{i=1}^{\tilde{n}}$, we can get an estimate $\hat{c}(\btheta)$ of the cost function. Then, the cost-aware version of SBI methods can be implemented as per \Cref{alg:ca-sbi}. Note that these initial simulations are not wasted since they can be recycled for SBI.

\paragraph{Neural SBI.} In the case of NPE/NLE, we need to estimate the losses in Equations \ref{eq:npe_loss} and \ref{eq:nle_loss} respectively. This requires approximating nested expectations with respect to the prior $p$, which can be expensive when $p$ places significant mass in regions of $\Theta$ where the simulator is expensive. We therefore use a cost-aware importance sampling: given samples $({\theta}_1, \dots, {\theta}_n) \sim \tilde{p}_g$ obtained using rejection sampling and the corresponding simulated data $x_{i1},\ldots,x_{im} \sim p(\x|\btheta_i)$, we can estimate the \textit{cost-aware NPE (Ca-NPE)} and \textit{cost-aware NLE (Ca-NLE)} losses as  
\begin{align*}
\hat \ell_{\text{Ca-NPE}}(\bphi) & =  -\frac{1}{m n} \sum_{j=1}^m \sum_{i=1}^n w_{\mathrm{Ca}}(\theta_i) \log q_{\phi} (\btheta_i \cond x_{ij})\\
\hat \ell_{\text{Ca-NLE}}(\bphi) & = -\frac{1}{m n} \sum_{j=1}^m \sum_{i=1}^n w_{\mathrm{Ca}}(\theta_i) \log q_\phi(x_{ij} \cond \theta_i).
\end{align*}
 
\paragraph{Approximate Bayesian computation.} The target is now the ABC approximate posterior $p_{\text{ABC}}(\btheta \cond \xo) \propto \mathbb{E}_{\x\sim p(\x \cond \btheta)}[1_{\{\varrho(\xo,\x) \leq \epsilon\}}] p(\theta)$ and we use a proposal 
$$\tilde p_{\text{ABC}}(\theta \cond x_o) \propto \mathbb{E}_{\x\sim p(\x \cond \btheta)}[1_{\{\varrho(\xo,\x) \leq \epsilon\}}]  \tilde{p}_g(\theta), $$ 
where $\tilde{p}_g(\theta) \propto p(\theta)/g(c(\theta))$.
The \emph{cost-aware ABC} algorithm therefore consists of sampling parameter values from $\tilde{p}_g$ through rejection sampling, using the accept/reject mechanism of ABC, and returning all accepted samples $(\theta_1, \dots, \theta_{n_\epsilon})$ weighted by $w_{\mathrm{Ca}}(\theta_i) = w(\theta_i)/\sum_{i=1}^{n_\epsilon} w(\theta_i)$. The target ABC posterior $ p_{\text{ABC}}(\btheta \cond \xo)$ can then be approximated as $\sum_{i=1}^{n_\epsilon} w_{\text{Ca}}({\theta}_i) \delta_{{\theta}_i}$, where $\delta_{{\theta}_i}$ is a Dirac delta mass at ${\theta}_i$.
Note that this is a consistent estimator despite the fact that the normalisation constants of $\tilde{p}_g(\btheta)$, $\tilde{p}_\text{ABC}(\btheta \cond \x_o)$ and $p(\btheta \cond \x_o)$ are all unknown. 

\begin{figure}
  \begin{center}
    \includegraphics[width=0.235\textwidth]{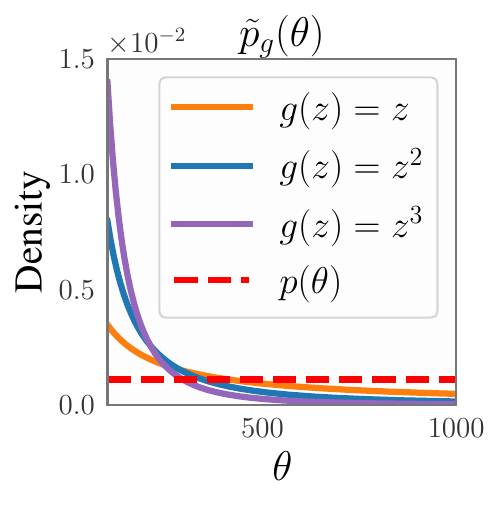}
    \includegraphics[width=0.235\textwidth]{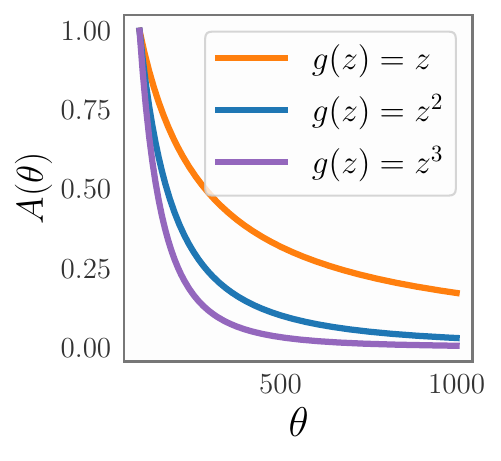}
  \end{center}
  \vspace{-2mm}
  \caption{\textit{Left:} Cost-aware prior $\tilde p_g(\theta)$ for different penalty functions $g(z) = z^k$ using prior $\mathcal{U}(10^2,10^3)$ for the Gamma experiment. The cost increases linearly with $\theta$, see \Cref{fig:gamma_experiment_npe}(a). \textit{Right:} Acceptance probability $A(\theta)$ as a function of $\theta$ for different $g$ functions.}
  \label{fig:gamma_priors}
\end{figure}
\section{EXPERIMENTS}
\label{sec:experiments}
We will now demonstrate the performance of our cost-aware versions of ABC, NPE, and NLE against their standard  counterparts. We use the $\mathtt{sbi}$ library \citep{Tejero-Cantero2020} for NPE and NLE. We use penalties of the form $g(z) = z^k$, and present results with different choices of $k$. We also present results for multiple importance sampling with $J=4$ components with equal proportions, which includes the target $p$. 
The code is available at \url{https://github.com/huangdaolang/cost-aware-sbi}.

\begin{figure*}
    \centering
    \subfigure[]{\includegraphics[width=0.22\linewidth]{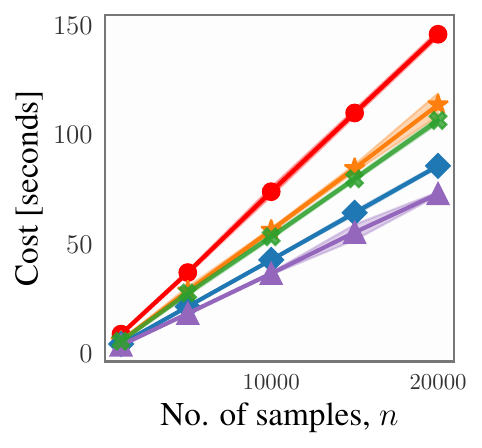}}
    \subfigure[]{\includegraphics[width=0.211\linewidth]{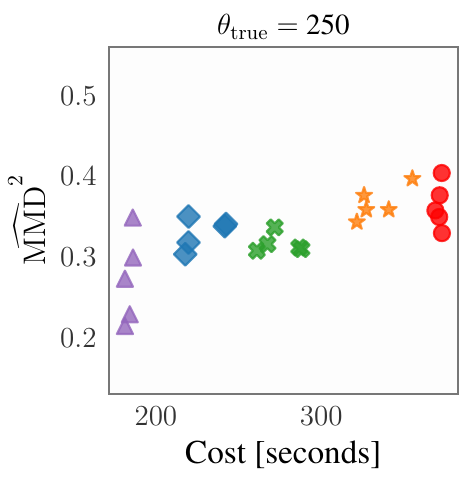}}
    \subfigure[]{\includegraphics[width=0.2\linewidth]{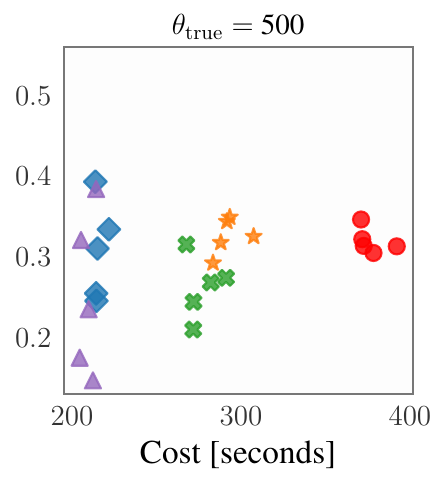}}
    \subfigure[]{\includegraphics[trim={7 0 0 0}, clip, width=0.185\linewidth]{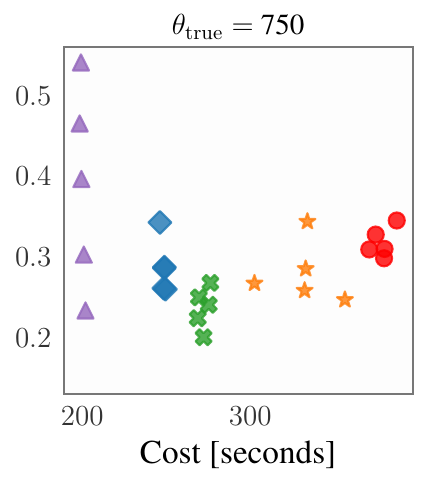}}
    \hspace{-2ex}
    \includegraphics[trim={60 260 30 0}, clip, width=0.16\linewidth]{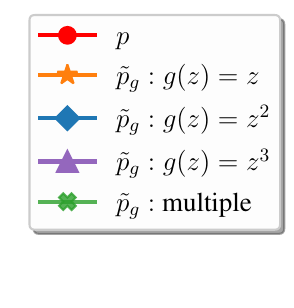}
    \vspace{-1ex}
    \caption{\textbf{The Gamma experiment.} (a) Cost of simulating $n$ data-points, each with $m=500$ Gamma samples, using the prior and different cost-aware proposals. (b)-(d) MMD between the ABC posteriors and the true posterior for different values of $\theta_{\mathrm{true}}$ over five independent runs with $n=50,000$ and $\epsilon = 0.05$. Sample mean and standard deviation of $m$ points are taken as statistics.  The corresponding NPE plots are shown in \Cref{app:gamma_additional}.}
    \label{fig:gamma_experiment}
\end{figure*}
\subsection{Illustrative Example: Gamma simulator}
We begin with an illustrative example which does not require SBI since it has a tractable likelihood, but which allows us to assess cost-aware SBI in detail. When simulating from a $\text{Gamma}(\theta, 1)$ using the popular Ahrens-Dieter acceptance-rejection method \citep{Ahrens1982}, the number of uniform draws required increases with the shape parameter $\theta$. The computational cost scales linearly with $\theta$, and we found that $c(100) \approx 0.002 \text{ seconds}$ and $c(1000) \approx 0.02 \text{ seconds}$ for $m=500$ samples on our machine. Taking $p$ to be uniform on $\Theta=[10^2,10^3]$, we can obtain  $\tilde{p}_g$ for $g(z) = z^k$, $k=1,2,3$ in closed-form; see \Cref{fig:gamma_priors} (left). As expected, the cost-aware proposals place less mass on the computationally expensive regions of $\Theta$ (i.e. larger values of $\theta$) and the extent of the penalisation is controlled by $k$. \Cref{fig:gamma_priors} (right) shows the acceptance probabilities $A(\theta)$ used to sample from $\tilde{p}_g$ via rejection sampling, which naturally reduces as the cost increases. This is not a problem since the cost of simulating from $p$ is very small (of the order of $10^{-7} \text{ s}$).

\begin{table}
\centering
\caption{Cost-efficiency trade-off for Gamma$(\theta, 1)$.}
\vspace{-1ex}
\scalebox{0.75}{
\begin{tabular}{@{}rclcccc@{}}
\toprule
\multicolumn{1}{l}{} & SBI   &  & \multicolumn{4}{c}{Ca-SBI}                              \\ \cmidrule(lr){2-2} \cmidrule(l){4-7} 
$g(z)$               & -     &  & $z^{0.5}$ & $z$ & $z^2$ & $z^3$ \\ \midrule
ESS         & 1.0   &  & 0.94  & 0.80  & 0.42   & 0.16\\
CG          & 1.0   &  & 1.11  & 1.25  & 1.62  & 2.06\\
 Cost SBI$/$Cost Ca-SBI  & 1.0   &  & 1.13  & 1.26  & 1.65  & 2.0\\
 \midrule
CG$\times$ESS             & $1.0$ &  & $1.05$      & $1.0$        & $0.68$         & $0.33$ \\ \bottomrule
\end{tabular}
}
\label{table:ess_cost}
\end{table}

To choose $g$, we analyse the cost-efficiency trade-off by simulating $n=20,000$ samples from the Gamma simulator in \Cref{table:ess_cost}. First, we see that CG is very close to the ratio of the total cost of simulations using SBI and Ca-SBI for different $g$, indicating that CG is a good proxy for measuring the cost benefits. Unsurprisingly, ESS decreases as $k$ increases whilst the cost increases when $k$ increases. This confirms that CG$\times$ESS is a reasonable metric for balancing efficiency and cost when selecting $g$. Taking $k=1$ yields a value of $1.0$, similar to using the prior $p$ for sampling. We therefore take the prior and cost-aware proposals with $k=\{1, 2, 3\}$ as the components for multiple importance sampling.

We plot the cost of simulating $m=500$ data points from the Gamma simulator for different sampling distributions as a function of number of parameter samples $n$ in \Cref{fig:gamma_experiment}(a). Naturally, the cost reduces as the exponent $k$ increases, with $k=2$ and $k=3$ taking less than half the time compared to using the prior. In \Cref{fig:gamma_experiment}(b)-(d), we measure the cost-versus-efficiency trade-off using the maximum mean discrepancy (MMD) \citep{Gretton2012JMLR} between a reference posterior (obtained using the $\mathtt{pymc3}$ library \citep{pymc3_2016}) and the ABC posteriors as a function of cost for different true parameter values. The corresponding plots for NPE are shown in \Cref{app:gamma_additional}.
As the penalty on the cost increases (by increasing $k$), the posterior approximation improves
when the true value lies in the low-cost region ($\theta_{\mathrm{true}} = 250$) but can degrade slightly otherwise (see $\theta_{\mathrm{true}} = 750$).
The multiple importance sampling approach consistently achieves similar posterior accuracy as using the prior (i.e. classical rejection ABC), with the added benefit of around 25\% reduction in total simulation cost.

\begin{table*}[h]
\centering
\caption{\text{NPE and Ca-NPE on three SIR models.} The mean and \textcolor{gray}{standard deviation} from $50$ runs are reported. Ca-NPE reaches comparable performance to NPE when $k$ is small whilst taking significantly less time than NPE.}
\scalebox{0.85}{
\begin{tabular}{rccccclcccc}
\hline
                & \multicolumn{5}{c}{$\widehat{\text{MMD}}^2$($\downarrow$)}                                                                                                                                                                      & \multicolumn{1}{c}{} & \multicolumn{4}{c}{Time saved ($\uparrow$)}                                                                                                                                                   \\ \cline{2-6} \cline{8-11} 
                & \multicolumn{1}{l}{}            & \multicolumn{1}{l}{}                                              & \multicolumn{1}{l}{}                                      & \multicolumn{1}{l}{}                                          &                      & \multicolumn{1}{l}{}                                              & \multicolumn{1}{l}{}                                      & \multicolumn{1}{l}{}     & \multicolumn{1}{l}{} & \multicolumn{1}{l}{}                                     \\
                & NPE                             & \begin{tabular}[c]{@{}c@{}}Ca-NPE\\ $g(z) = z^{0.5}$\end{tabular} & \begin{tabular}[c]{@{}c@{}}Ca-NPE\\ $g(z)=z$\end{tabular} & \begin{tabular}[c]{@{}c@{}}Ca-NPE\\ $g(z) = z^2$\end{tabular} & \begin{tabular}[c]{@{}c@{}}Ca-NPE\\ multiple\end{tabular} & \multicolumn{1}{c}{} & \begin{tabular}[c]{@{}c@{}}Ca-NPE\\ $g(z) = z^{0.5}$\end{tabular} & \begin{tabular}[c]{@{}c@{}}Ca-NPE\\ $g(z)=z$\end{tabular} & \begin{tabular}[c]{@{}c@{}}Ca-NPE\\ $g(z) = z^2$\end{tabular} & \begin{tabular}[c]{@{}c@{}}Ca-NPE\\ multiple\end{tabular} \\ \hline
Homogen. & $0.02 \textcolor{gray}{(0.02)}$ & $0.02 \textcolor{gray}{(0.01)}$                                                                  & $0.02 \textcolor{gray}{(0.02)}$                                                          & $0.23 \textcolor{gray}{(0.08)}$                                    &   $ 0.05 \textcolor{gray}{(0.04)}$  &  & $16\% \textcolor{gray}{(2)}$                      & $38\% \textcolor{gray}{(2)}$                     & $70\% \textcolor{gray}{(2)}$ & $30\% \textcolor{gray}{(5)}$ \\
Temporal    & $0.03 \textcolor{gray}{(0.03)}$ & $0.06 \textcolor{gray}{(0.03)}$                                   & $0.07 \textcolor{gray}{(0.03)}$ & $0.07 \textcolor{gray}{(0.03)}$                                                                          &     $0.05 \textcolor{gray}{(0.04)}$     &                      & $36\% \textcolor{gray}{(4)}$                                      & $65\% \textcolor{gray}{(2)}$                              & $85\% \textcolor{gray}{(1)}$     & $24\% \textcolor{gray}{(5)}$                                                        \\
Bernoulli   & $0.02 \textcolor{gray}{(0.00)}$                                & $0.02 \textcolor{gray}{(0.00)}$                                                                  & $0.02 \textcolor{gray}{(0.01)}$                                                          & $0.04 \textcolor{gray}{(0.01)}$                                                              &  $ 0.02 \textcolor{gray}{(0.00)}$   &                 & $23\% \textcolor{gray}{(4)}$                                                                  & $37\% \textcolor{gray}{(4)}$                                                          & $47\% \textcolor{gray}{(3)}$       & $25\% \textcolor{gray}{(6)}$                                                                            \\ \hline
\end{tabular}
}
\label{tab:sir}
\end{table*}

\begin{figure*}[t]
    \centering
    \includegraphics[width=1\textwidth]{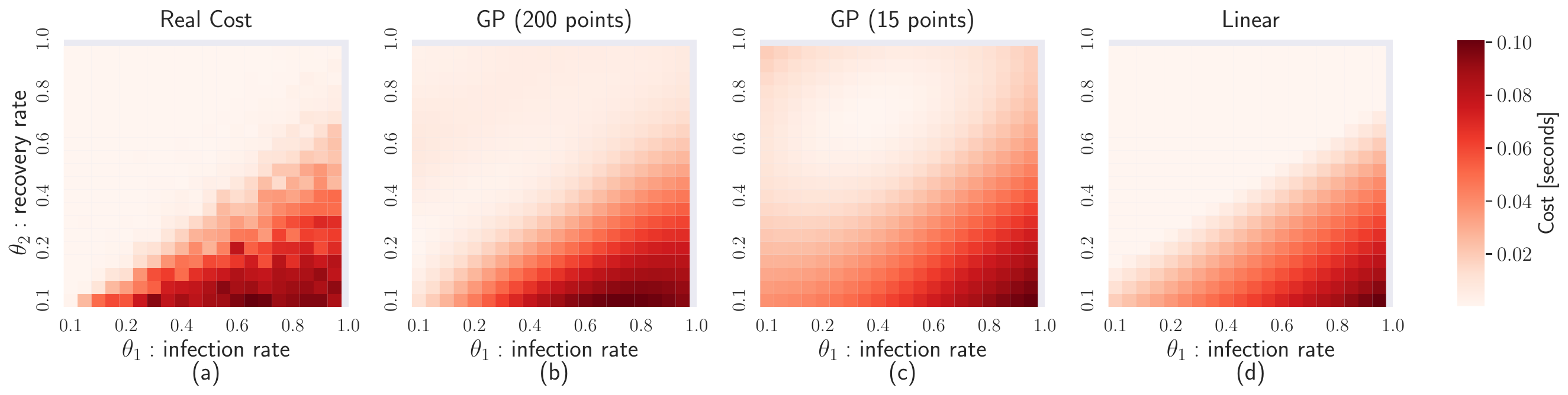}
    \caption{Cost function estimate of the temporal SIR model at varying levels of accuracy. (a) The real cost function estimated using a $20\times20$ grid with 50 samples for each grid (same as \Cref{fig:intro}). (b) Estimated cost using a GP model trained on 200 data points. (c) Estimated cost using a GP model trained on only 15 data points. (d) Estimated cost using a linear model trained on 200 data points. }
    \label{fig:cost_estimate_comparison}
\end{figure*}

\subsection{Epidemiology models}\label{sec:experiments_epidemiology}
We now consider variants of the SIR model in epidemiology, where SBI methods have been extensively applied \citep{mckinley2009inference, neal2012efficient, mckinley2014simulation, Kypraios2017, McKinley2018}. As discussed in \Cref{sec:introduction}, the cost for these models can vary significantly depending on parameters such as infection and recovery rates. We consider three variants: the homogeneous, temporal, and Bernoulli SIR. These models have between $1$ and $3$ parameters impacting the cost.
Detailed descriptions and their corresponding estimated cost functions are in \Cref{app:sir_description}. 

We compare the performance in terms of MMD between a reference NPE posterior trained on $n=50,000$ samples and the NPE (or Ca-NPE) posteriors trained on a smaller set of $n=5,000$ samples drawn from $p(\theta)$ or our $\tilde{p}_g(\theta)$ with different choice of $g$. We compute the time saved by Ca-NPE as 1 minus the ratio of the total simulation time using $\tilde{p}_g$ to that of using $p$. We report numbers for penalty function with $k=\{0.5, 1, 2\}$ and the multiple proposal in \Cref{tab:sir}. Our Ca-NPE method reduces simulation time across different SIR models, whilst maintaining comparable accuracy to the standard NPE approach. For instance, Ca-NPE with $g(z)=z$ achieves a 37\% reduction (2.4 hrs) in simulation cost without sacrificing performance in the Bernoulli SIR model. Similarly, for the temporal SIR model, $k=2$ provides the most significant time savings of 85\% (380 s), with only a slight increase in MMD. Similar results are observed when using the C2ST and marginal two-sample KS tests metrics; see \Cref{app:sir_description}.

We now also study the impact of the accuracy of the cost function estimator using the temporal SIR model. Apart from using a GP to estimate the cost function with 200 data-points, we also estimate the cost using a GP trained on just 15 points, and with a linear model trained on 200 points. The corresponding cost plots are shown in \Cref{fig:cost_estimate_comparison}, along with the real cost function (same as \Cref{fig:intro}). We then use these different cost estimates to run Ca-NPE with different penalty function, and report the results in \Cref{tab:varying_cost}.
We observe that the accuracy of the cost estimate does not impact the accuracy of the posterior estimation, as the MMD values for all the cases are similar. Moreover, even if the cost function is not accurately estimated, we are still able to reduce the computational cost of doing SBI through a cost-aware approach; see for example the GP(15 points) and linear cost models. Hence, running our cost-aware method with rough estimates of the cost function also provides computational gain, without losing out on posterior accuracy.

\begin{table*}[t]
\centering
\caption{Effect of inaccurate cost function estimation on Ca-SBI. MMD ($\downarrow$) and time saved ($\uparrow$) for Ca-NPE on the temporal SIR model with different cost models. The mean and \textcolor{gray}{standard deviation} from $50$ runs are reported.}
\scalebox{0.85}{
\begin{tabular}{@{}rcclcclcclcc@{}}
\toprule
\multicolumn{1}{c}{\multirow{2}{*}{Cost model}} & \multicolumn{2}{c}{Ca-NPE $(g(z) = z^{0.5})$}                     &  & \multicolumn{2}{c}{Ca-NPE $(g(z) = z)$}                        &  & \multicolumn{2}{c}{Ca-NPE $(g(z) = z^{2})$}                     &  & \multicolumn{2}{c}{Ca-NPE (multiple)}                           \\ \cmidrule(lr){2-3} \cmidrule(lr){5-6} \cmidrule(lr){8-9} \cmidrule(l){11-12} 
\multicolumn{1}{c}{}                            & $\widehat{\text{MMD}}^2$                              & Time saved                     &  & $\widehat{\text{MMD}}^2$                             & Time saved                   &  & $\widehat{\text{MMD}}^2$                             & Time saved                    &  & $\widehat{\text{MMD}}^2$                             & Time saved                    \\ \midrule
GP (200 points)                                 & $ 0.06 \textcolor{gray}{(0.03)}$ & $ 36 \% \textcolor{gray}{(4)}$ &  & $0.07 \textcolor{gray}{(0.03)}$ & $65\% \textcolor{gray}{(2)}$ &  & $0.07 \textcolor{gray}{(0.03)}$ & $85\% \textcolor{gray}{(1)}$  &  & $0.05 \textcolor{gray}{(0.04)}$ & $ 24\% \textcolor{gray}{(5)}$ \\
GP (15 points)                                  & $0.06 \textcolor{gray}{(0.03)}$  & $8\% \textcolor{gray}{(3)}$    &  & $0.06 \textcolor{gray}{(0.03)}$ & $33\% \textcolor{gray}{(2)}$ &  & $0.08 \textcolor{gray}{(0.03)}$ & $69 \% \textcolor{gray}{(1)}$ &  & $0.05 \textcolor{gray}{(0.04)}$ & $15\% \textcolor{gray}{(3)}$  \\
Linear                                          & $ 0.07 \textcolor{gray}{(0.02)}$ & $19\% \textcolor{gray}{(5)}$   &  & $0.06 \textcolor{gray}{(0.02)}$ & $34\% \textcolor{gray}{(6)}$ &  & $0.06 \textcolor{gray}{(0.02)}$ & $61\%\textcolor{gray}{(2)}$   &  & $0.07 \textcolor{gray}{(0.02)}$ & $27\% \textcolor{gray}{(3)}$  \\ \bottomrule
\end{tabular}
}
\label{tab:varying_cost}
\end{table*}

\begin{figure*}[t]
    \centering
    \includegraphics[width=1\textwidth]{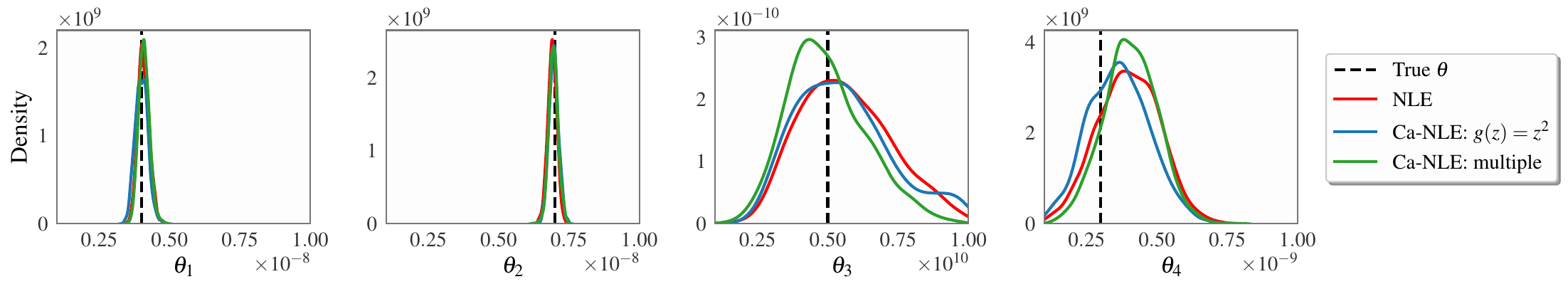}
    \caption{Marginals of the approximate posterior for the radio propagation model using NLE and our Ca-NLE method with $n=10,000$. The Ca-NLE methods perform similar to NLE, whilst saving hours of computation. }
    \label{fig:turin_nle}
\end{figure*}

\subsection{Radio propagation model}
\label{sec:radio_experiment}

Finally, we apply our method on a computationally expensive real-world simulator with four parameters from the radio propagation field; see \citep{Bharti22a, Huang2023}. Driven by an underlying point process, this model simulates an 801-dimensional complex-valued time series data which represents a radio signal. The cost of simulation increases linearly with the arrival rate of the point process, which is one of the four parameters. The other three parameters, governing the amplitude and noise present in the signal, do not affect the cost. For each parameter $\theta$, we simulate $m=50$ iid time series and use six summary statistics based on temporal moments \citep{Bharti2022} of the signals, as described in \Cref{app:radio_propagation}.

We generate $n=10,000$ realisations using a uniform prior, which took $15.6$ hrs. In contrast, simulating the same number of samples from our cost-aware proposal took $8.8$ hrs with $g(z) = z^2$ and $11.6$ hrs with multiple importance sampling ($k=\{1,2,3\}$), leading to a cost reduction of 44\% and 26\%, respectively. All three posterior distributions are very similar, with multiple importance sampling yielding the most concentrated posterior around the true value for the arrival rate $\theta_3$. Therefore, using our approach, we save hours of computational time without any degradation in performance. We plot the corresponding NLE posteriors in \Cref{fig:turin_nle} and observe similar results.

Interestingly,  Ca-SBI is embarrassingly parallelisable and can therefore lead to futher computational gains if the right resources for parallelisation are available to the user. To demonstrate this, we generated $n=10,000$ realisations again from the radio propagation model by parallelising computation with 50 cores for both standard SBI and Ca-SBI with $g(z) = z^2$. As expected, the total run time reduced by a factor of approximately $1/50$, with standard SBI taking $21$ mins (instead of $15.6$ hrs) and Ca-SBI taking $10.6$ mins (instead of 8.8 hrs), resulting in a computational speed-up of around $49.5\%$, similar to what we observed before parallelisation.

\section{CONCLUSION}
\label{sec:conclusion}

We proposed a family of cost-aware importance sampling distributions which can reduce the overall cost when downstream tasks are expensive and have a cost which varies over the sampling space. For SBI methods, this is particularly helpful when the cost of simulating from the model varies across parameters. As our method only depends on the prior, we could show that it can be applied to the most popular SBI methods including ABC, NPE, and NLE. Other methods such as regression-adjustment ABC \citep{Beaumont2002}, generalised Bayesian inference \citep{Pacchiardi2021}, Bayesian synthetic likelihood \citep{price2018bayesian,Frazier2023}, and neural ratio estimation \citep{Durkan2020, Hermans2020} could also be made cost-aware in a similar manner.

\paragraph{Limitations and future work.} One interesting extension of the work could be to use cost-aware sampling beyond SBI. As importance sampling suffers in high dimensions, our method may not work as well if there are many parameters that affect the cost function. The performance also degrades when the true value lies in the high-cost region. However, these issues could be alleviated by using adaptive importance sampling \citep{cornuet2012adaptive, martino2015adaptive} where the penalty function is adapted based on an initial coarse estimate of where the posterior lies. For SBI, a limitation of our approach is that it does not apply to optimisation-based SBI methods such as minimum distance estimation \citep{Briol2019MMD, Key2021, Dellaporta2022}, the method of simulated moments \citep{Hall2003}, or regression-based methods \citep{Gutmann2016}. Developing cost-aware versions of these methods is an interesting avenue for future work, and could be approached with cost-aware Bayesian optimisation as proposed in \citet{Lee2020}. 

\subsubsection*{Acknowledgements}
The authors are grateful to Art Owen and Dennis Prangle for pointing out relevant related work. AB, DH and SK were supported by the Research Council of Finland (Flagship programme: Finnish Center for Artificial Intelligence FCAI).
AB was also supported by the Research Council of Finland grant no. 362534.
SK was also supported by the UKRI Turing AI World-Leading Researcher Fellowship, [EP/W002973/1].
FXB was supported by the EPSRC grant [EP/Y022300/1].


\bibliography{bibliography}
\bibliographystyle{apalike}

\appendix
\onecolumn
{
\begin{center}
\Large
    \textbf{Supplementary Materials}
\end{center}
}

\Cref{app:proofs} contains the proofs of the theoretical results presented in the main text. In \Cref{app:additional_results}, we derive closed-form expressions for the cost-aware proposal and the computational gain (CG) in certain cases. \Cref{app:implementation} consists of the implementation details and additional results for the experiments conducted in \Cref{sec:experiments}.

\section{Proof of theoretical results}
\label{app:proofs}

\subsection{Proof of \Cref{prop:rejection_sampling}}
\label{app:proof_rejection}
\begin{proof}
    Consider the cost-aware proposal  $\proposal(\btheta) \propto \pi(\theta) / g(c(\btheta))$. In order to ensure that  $\proposal$ is indeed a probability density function, we need to show that (i) it is non-negative, and (ii) can be normalised; i.e. there exists a normalisation constant $B>0$ which is finite. For (i), we know that $\pi$ is a probability density function, and that $g_{\text{min}}>0$, therefore $\proposal$ is a ratio of non-negative quantities and must be non-negative. For (ii), we can once again use the existence of $g_{\text{min}}>0$ to guarantee the normalisation constant is finite since:
    \begin{align}\label{eq:B_upper_bound}
        B = \int_\Theta \frac{\pi(\btheta)}{g(c(\btheta))} d\btheta \leq \frac{\int_\Theta \pi(\btheta) d\btheta}{\inf_{\btheta \in \Theta} g(c(\btheta))} = \frac{1}{g_{\mathrm{min}}}.
    \end{align}
    Here, the inequality comes from the fact that $g \circ c$ is strictly positive, so we must have that $B < 1/g_{\mathrm{min}}  < \infty$. This concludes our proof that $\proposal$ is a valid probability density function. 
    
    To sample from proposal $\proposal$ using rejection sampling, we need to find a constant $M>0$ such that the condition $\proposal(\theta) \leq M \pi(\theta)$ holds. To that end,
    \begin{equation*}
        \frac{\proposal(\theta)}{\pi(\theta)} =  \frac{\pi(\theta)}{B g(c(\btheta)) \pi(\btheta)} 
        = \frac{1}{B  g(c(\btheta)) } \leq \frac{1}{B g_{\mathrm{min}}} = M,
    \end{equation*}
    where the first equality follows from the definition of $\proposal$, the second equality by cancellation of $\pi$ in the numerator and denominator, and the inequality by the fact that $g_{\text{min}}$ is the infimum of $g \circ c$.
    Note that $g \circ c$ is strictly positive, and therefore $M < \infty$.
    The acceptance probability $A(\btheta)$ for the rejection sampler can hence be written as 
\begin{equation*}
        A(\theta) = \frac{\proposal(\theta)}{M \pi(\theta)} = \frac{\pi(\theta) B g_{\mathrm{min}} }{B g(c(\btheta)) \pi(\btheta)} = \frac{g_{\mathrm{min}} }{g(c(\btheta))}, 
    \end{equation*}
     which completes the proof.
\end{proof}

\subsection{Proof of \Cref{prop:finite_variance}}
\label{app:proof_variance}

\begin{proof}
We start by proving almost-sure convergence. According to Theorem 9.2 of \citet[Ch. 9]{mcbook}, the self-normalised importance sampling estimator $\hat{\mu}_n^{\text{Ca}}$ converges to $\mu$ in probability as $n$ goes to infinity if the proposal $\proposal$ is a probability density function on $\Theta \subseteq \mathbb{R}^p$ and $\proposal(\theta) > 0$ whenever the target $\pi(\theta)> 0$. From \Cref{prop:rejection_sampling}, we know that $\proposal(\theta) = \pi(\theta) / Bg(c(\theta))$ is a valid density with $0< B< \infty$, and that $\inf_{\theta \in \Theta} g(c(\theta)) = g_{\mathrm{min}} > 0$. As a result, the denominator $Bg(c(\theta))$ is greater than zero for all $\theta \in \Theta$, and hence $\pi(\theta) > 0$ whenever $\proposal(\theta)>0$, which completes the proof of the first statement.

We now move on to proving the bounds on the weights. Recall that 
\begin{align*}
    w_{\mathrm{Ca}}(\theta_i) = \frac{w(\theta_i)}{\sum_{j=1}^n w(\theta_j)} = \frac{g(c(\theta_i))}{\sum_{j=1}^n g(c(\theta_j))}
    \leq \frac{\sup_{\theta \in \Theta} g(c(\theta))}{n \inf_{\theta \in \Theta} g(c(\theta))}  = \frac{g_{\text{max}}}{n g_{\text{min}}}.
\end{align*}
Similarly, we also have a lower bound by taking an infimum on the numerator and supremum on the denominator:
\begin{align*}
     w_{\mathrm{Ca}}(\theta_i) \geq \frac{g_{\text{min}}}{n g_{\text{max}}}.
\end{align*}
Finally, we prove the bounds on the variance $\sigma^2_{\text{Ca}}$. Since we are considering a self-normalised importance sampling estimator, the variance of $\hat{\mu}^{\text{Ca}}_n$ can be written as \citep[Ch. 9]{mcbook}:
    \begin{align}
    n \sigma^2_{\text{Ca}} 
    =  \int_{\Theta} \frac{\pi(\theta)^2 (f(\theta) - \mu)^2}{\proposal(\theta)} d\theta 
    & = \int_{\Theta} \frac{B g(c(\theta)) \pi(\theta)^2 (f(\theta) - \mu)^2}{\pi(\theta)} d\theta  = B \int_{\Theta} g(c(\theta)) \pi(\theta) (f(\theta) - \mu)^2 d\theta \label{eq:ca_var}
    \end{align}
    Upper bounding this expression, we get:
    \begin{align*}
    n \sigma^2_{\text{Ca}}  \leq B g_{\text{max}} \int_{\Theta}  \pi(\theta) (f(\theta) - \mu)^2  d\theta 
    & \leq \frac{g_{\text{max}}}{g_{\text{min}}} \int_{\Theta}  \pi(\theta) (f(\theta) - \mu)^2  d\theta \\
    & = \frac{g_{\text{max}}}{g_{\text{min}}} \int_{\Theta}  \pi(\theta) (f^2(\theta) - 2 f(\theta)\mu +\mu^2) d\theta 
    = \frac{g_{\text{max}}}{g_{\text{min}}} \left(n \sigma^2_{\text{MC}} - \mu^2 \right).
\end{align*}
To derive the above, we first used the definition of our cost-aware proposal, then used Hölder's inequality to take out the supremum of $g$ and the bound in \Cref{eq:B_upper_bound} to upper bound $B$. We then concluded by expanding the square and simplifying using the definitions of $\mu$ and $\sigma^2_{\text{MC}}$. Dividing by $n$ on both sides of the inequality gives our upper bound. 

Similarly, we can get the lower bound by using a lower bound on $B$ (rather than an upper bound). Such a bound can be obtained as:
\begin{align}\label{eq:B_lower_bound}
    B = \int_\Theta \frac{\pi(\btheta)}{g(c(\btheta))} d\btheta \geq \frac{\int_\Theta \pi(\btheta) d\btheta}{\sup_{\btheta \in \Theta} g(c(\btheta))} = \frac{1}{g_{\mathrm{max}}}.
\end{align}
Using \Cref{eq:B_lower_bound} to lower bound \Cref{eq:ca_var} and taking an infimum over $g$, we get
\begin{align*}
    n \sigma^2_{\text{Ca}} \geq  \frac{g_{\text{min}}}{g_{\text{max}}} \int_{\Theta}  \pi(\theta) (f(\theta) - \mu)^2  d\theta   =\frac{g_{\text{min}}}{g_{\text{max}}}   \left(n \sigma^2_{\text{MC}} - \mu^2 \right).
\end{align*}
Once again, dividing both sides by $n$ gives the required bound. This concludes our proof.
\end{proof}

\subsection{Proof of \Cref{prop:bound_ESS}}\label{app:proof_ESS_bounds}

\begin{proof}
We start with the upper bound on the ESS, and take a similar approach to that used to bound the weights in \Cref{prop:finite_variance}:
    \begin{align*}
        \text{ESS}  
        = \frac{\left( \sum_{i=1}^n g(c(\btheta_i)) \right)^2}{n\sum_{i=1}^n g(c(\btheta_i))^2}  
        \leq 
        \frac{\sup_{\theta \in \Theta} \left( \sum_{i=1}^n g(c(\btheta)) \right)^2}{ \inf_{\theta \in \Theta} n\sum_{i=1}^n g(c(\btheta))^2} 
        = \frac{n^2 g_{\text{max}}^2}{n^2 g_{\text{min}}^2} = \left(\frac{ g_{\text{max}}}{g_{\text{min}}}\right)^2.
    \end{align*}
    Similarly (but taking an infimum on the numerator and supremum on the denominator), we get a corresponding lower bound
    \begin{align*}
        \text{ESS} 
        = \frac{\left( \sum_{i=1}^n g(c(\btheta_i)) \right)^2}{n\sum_{i=1}^n g(c(\btheta_i))^2}  
        \geq  \frac{\inf_{\theta \in \Theta} \left( \sum_{i=1}^n g(c(\btheta)) \right)^2}{ \sup_{\theta \in \Theta} n\sum_{i=1}^n g(c(\btheta))^2}  
        = \left(\frac{ g_{\text{min}}}{g_{\text{max}}}\right)^2,
    \end{align*}
    which concludes our proof.
\end{proof}

\subsection{Proof of \Cref{prop:CG_bounds}}\label{app:proof_CG_bounds}

\begin{proof}
The fact that the CG is lower-bounded by $1$ is trivial and simply follows from the fact that $\proposal$ is explicitly putting less mass than $\pi$ in regions of higher cost. 

For the upper bound, we start by proving an explicit expression for the CG in terms of $B$ and $g(c(\theta))$:
\begin{align*}
    \text{CG} 
    = \frac{\int_\Theta c(\btheta) \pi(\btheta) d\btheta}{\int_\Theta c(\btheta) \proposal(\btheta) d\btheta} 
    = \frac{\int_\Theta c(\btheta) \pi(\btheta) d\btheta}{\int_\Theta c(\btheta) \frac{\pi(\btheta)}{B g(c(\theta))} d\btheta} 
    = B \frac{\int_\Theta c(\btheta) \pi(\btheta) d\btheta}{\int_\Theta c(\btheta) \frac{\pi(\btheta)}{g(c(\theta))} d\btheta} 
\end{align*}
Using the above, we get
\begin{align*}
     \text{CG} \leq \frac{B}{\inf_{\theta \in \Theta} \frac{1}{g(c(\theta))}} \left( \frac{\int_\Theta c(\btheta) \pi(\btheta) d\btheta}{ \int_\Theta c(\btheta) \pi(\btheta) d\btheta}\right)  = B g_{\text{max}} \leq \frac{g_{\text{max}}}{g_{\text{min}}}.
\end{align*}
where the last inequality follows from the fact that $B \leq \frac{1}{g_{\text{min}}}$, as established in \Cref{eq:B_upper_bound}.
\end{proof}

\section{Closed-form expressions of the cost-aware proposal and related quantities}\label{app:additional_results}
\subsection{Rejection sampling with closed-form proposal} \label{sec:app_closedforms}

In this section, we derive closed-form expressions for the cost-aware proposal in order to plot its density for different penalty functions in \Cref{fig:gamma_priors}.
Instead of sampling from the prior $p(\btheta)$ (taken to be a $\mathcal{U}(\theta_{\text{min}},\theta_{\text{max}})$ with $0<\theta_{\text{min}} < \theta_{\text{max}} < \infty$), we propose to generate parameter values from the cost-aware proposal distribution $\tilde{p}_g= \frac{p(\btheta)}{g(c(\btheta)) \times B}$:
where $B = \int_\Theta \frac{p(\btheta)}{g(c(\btheta))} d\btheta < \infty$ is the normalisation constant that ensures $\tilde p_g$ is a density. Although our method does not require $B$ to be known, it can be computed in closed-form for certain choices of the cost function $c(\theta)$, penalty function $g$, and the prior $p$ as given in \Cref{tab:B_expression}.
\begin{table}[h]
\centering
\caption{Closed-form expressions for the normalisation constant $B$ of cost-aware densities.}
\label{tab:B_expression}
\begin{tabular}{@{}ccl@{}}
\toprule
$c(\theta)$             & $g(z)$       & $B$ \\ \midrule

$\alpha \theta + \beta$ & $z^k, k>1$  & $\frac{1}{(\theta_{\mathrm{max}} - \theta_{\mathrm{min}})}\left[ \frac{(\alpha \theta + \beta)^{1-k}}{\alpha(1-k)} \right]_{\theta_{\mathrm{min}}}^{\theta_{\mathrm{max}}} $   \\

$\alpha \theta + \beta$                        & $z$        & $\frac{1}{(\theta_{\mathrm{max}} - \theta_{\mathrm{min}})}\left[ \frac{\log (\alpha \theta + \beta)}{\alpha} \right]_{\theta_{\mathrm{min}}}^{\theta_{\mathrm{max}}} $ \\

$\alpha \theta^2$                        & $1$              & $\frac{1}{(\theta_{\mathrm{max}} - \theta_{\mathrm{min}})}\left[ - \frac{1}{\alpha^2 \theta}\right]_{\theta_{\mathrm{min}}}^{\theta_{\mathrm{max}}} $

\\ \bottomrule
\end{tabular}
\end{table}

\subsection{Closed-form expressions for computational gain (CG)}
\label{app:CG_derivation}

\begin{figure}[H]
\caption{CG vs. $\btheta_{\mathrm{max}}$ for different choices of penalty function $g$ under uniform prior $\mathcal{U}(1, \btheta_{\mathrm{max}})$. }
  \begin{center}
    \includegraphics[width=0.45\textwidth]{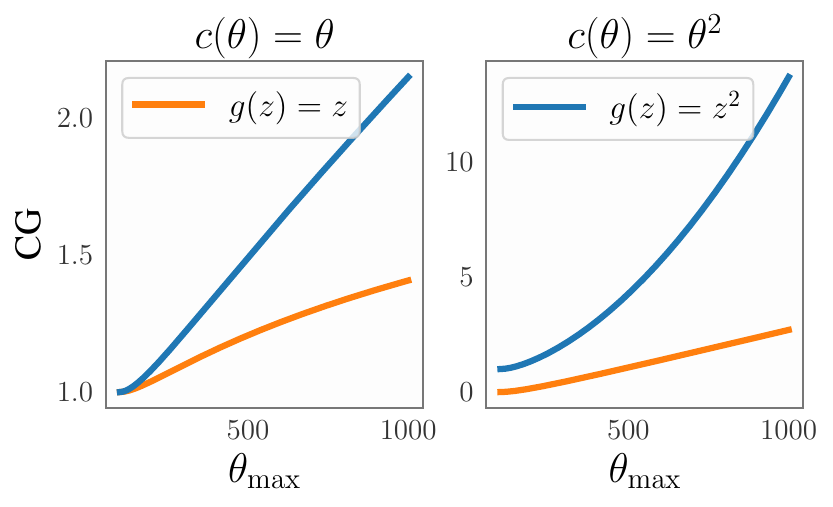}
  \end{center}
  
  \label{fig:CG}
\end{figure}

In case of a uniform prior distribution $\mathcal{U}(\btheta_{\mathrm{min}}, \btheta_{\mathrm{max}})$, the computational gain (CG) can be computed in closed-form for linear and quadratic choices of cost and penalty function. We report the CG expressions in \Cref{tab:CG}, and plot the CG as a function of $\theta_{\mathrm{max}}$ for linear and quadratic cost functions in \Cref{fig:CG}. Naturally, the gains are more significant when the cost is quadratic and $g(z)=z^2$.

\begin{multicols}{2}
\begin{table}[H]
    \centering
    \caption{Closed-form expressions of  computational gain for uniform prior $\mathcal{U}(\btheta_{\mathrm{min}}, \btheta_{\mathrm{max}})$.}
    \scalebox{1.05}{
    \begin{tabular}{ccc} \toprule
        $c(\btheta)$ & $g(z)$  & CG \\ \midrule
         $\alpha \btheta$ & $z$  & $\frac{(\btheta_{\mathrm{max}} + \btheta_{\mathrm{min}}) \log\left(\frac{\btheta_{\mathrm{max}}}{\btheta_{\mathrm{min}}}\right)}{2 (\btheta_{\mathrm{max}} - \btheta_{\mathrm{min}})}$ \\ \\
         $\alpha \btheta^2$ & $z$  & $\frac{(\btheta_{\mathrm{max}}^3 - \btheta_{\mathrm{min}}^3)}{3 (\btheta_{\mathrm{max}} - \btheta_{\mathrm{min}}) \btheta_{\mathrm{min}} \btheta_{\mathrm{max}}}$ \\ \\
         $\alpha \btheta$ & $z^2$  & $\frac{(\btheta_{\mathrm{max}}^2 - \btheta_{\mathrm{min}}^2)}{2 \btheta_{\mathrm{max}} \btheta_{\mathrm{min}} \log(\btheta_{\mathrm{max}} / \btheta_{\mathrm{min}})}$ \\ \\
         $\alpha \btheta^2$ & $z^2$  & $\frac{(\btheta_{\mathrm{max}}^3 - \btheta_{\mathrm{min}}^3)^2}{9 (\btheta_{\mathrm{max}} - \btheta_{\mathrm{min}})^2 \btheta_{\mathrm{max}}^2 \btheta_{\mathrm{min}}^2}$ \\ \bottomrule
    \end{tabular}%
    }
    \label{tab:CG}
\end{table}
\end{multicols}
We now present the derivation of the closed-form expressions of CG from \Cref{tab:CG}:
    \paragraph{\textbf{Case 1:}} $c(\btheta) = \alpha \btheta$, $g(z) = z$. We start by computing the expected  cost of the cost-aware method:
    \begin{align*}
        \int_{\Theta} c(\theta) \tilde{p}_g(\theta) d\theta =  \int_\Theta \frac{c(\btheta)  p(\btheta) }{B g(c(\btheta))} d\btheta =  \frac{1}{B} =  \left(\int_{\Theta} \frac{p(\btheta) }{g(c(\btheta))} d \btheta \right)^{-1} =  \left(\int_{\Theta} \frac{p(\btheta) }{c(\btheta)}d \btheta \right)^{-1} 
    \end{align*}
    \begin{align*}
        \text{CG} = \left(\int_{\Theta} c(\btheta) p(\btheta) d\btheta \right) 
        \left(\int_{\Theta} \frac{p(\btheta) }{c(\btheta)}d \btheta \right)
        &= \left(\frac{\alpha}{(\btheta_{\mathrm{max}} - \btheta_{\mathrm{min}})} \int_{\Theta} \btheta d \btheta \right)
        \left(\frac{1}{\alpha (\btheta_{\mathrm{max}} - \btheta_{\mathrm{min}})} \int_{\Theta} \frac{1}{\btheta} d\btheta \right)\\
        &= \frac{(\btheta_{\mathrm{max}} + \btheta_{\mathrm{min}}) \log(\frac{\btheta_{\mathrm{max}}}{\btheta_{\mathrm{min}}})}{2 (\btheta_{\mathrm{max}} - \btheta_{\mathrm{min}})}
    \end{align*}
    \paragraph{\textbf{Case 2:}} $c(\btheta) = \alpha \btheta^2$, $g(z) = z$
    \begin{align*}
        \text{CG} &= 
        \left(\frac{\alpha}{(\btheta_{\mathrm{max}} - \btheta_{\mathrm{min}})} \int_{\Theta} \btheta^2 d \btheta \right)
        \left( \frac{1}{\alpha (\btheta_{\mathrm{max}} - \btheta_{\mathrm{min}})} \int_{\Theta} \frac{1}{\btheta^2} d\btheta \right)
        \\
        &= \left(\frac{1}{(\btheta_{\mathrm{max}} - \btheta_{\mathrm{min}})^2} \left[ \frac{\btheta^3}{3} \right]_{\btheta_{\mathrm{min}}}^{\btheta_{\mathrm{max}}}\right)
        \left[- \frac{1}{\btheta} \right]_{\btheta_{\mathrm{min}}}^{\btheta_{\mathrm{max}}} = \frac{(\btheta_{\mathrm{max}}^3 - \btheta_{\mathrm{min}}^3)}{3 (\btheta_{\mathrm{max}} - \btheta_{\mathrm{min}}) \btheta_{\mathrm{min}} \btheta_{\mathrm{max}}}
    \end{align*}
    \paragraph{\textbf{Case 3:}} $c(\btheta) = \alpha \btheta$, $g(z) = z^2$
    \begin{align*}
        \int_\Theta c(\theta) \tilde{p}_g(\theta) d\theta = \frac{1}{B} \int_\Theta \frac{c(\btheta)  p(\btheta)}{ c(\btheta)^2}  d\btheta =   \left(\int_\Theta \frac{p(\btheta)}{c(\btheta)} d\btheta \right) \left( \int \frac{p(\btheta)}{c(\btheta)^2}  d\btheta \right)^{-1} 
    \end{align*}
    \begin{align*}
        \text{CG} &= \left(\frac{\int_{\Theta} c(\btheta) p(\btheta) d\btheta}{\int_\Theta \frac{p(\btheta)}{ c(\btheta)}  d\btheta}\right) 
        \left(\int_{\Theta} \frac{p(\btheta)}{c(\btheta)^2}  d\btheta\right)  \\
        &= \left(\frac{\alpha \left[\btheta^2\right]_{\btheta_{\mathrm{min}}}^{\btheta_{\mathrm{max}}}}{2(\btheta_{\mathrm{max}} - \btheta_{\mathrm{min}})}\right) 
        \left(\frac{1}{\alpha^2 (\btheta_{\mathrm{max}} - \btheta_{\mathrm{min}})} \left[-\frac{1}{\btheta} \right]_{\btheta_{\mathrm{min}}}^{\btheta_{\mathrm{max}}}\right) 
        \left( \frac{1}{\frac{1}{\alpha (\btheta_{\mathrm{max}} - \btheta_{\mathrm{min}})} \left[\log \btheta \right]_{\btheta_{\mathrm{min}}}^{\btheta_{\mathrm{max}}}} \right)\\
        &= \frac{(\btheta_{\mathrm{max}} + \btheta_{\mathrm{min}})}{2 (\log \btheta_{\mathrm{max}} - \log \btheta_{\mathrm{min}})} \left(\frac{1}{\btheta_{\mathrm{min}}} - \frac{1}{\btheta_{\mathrm{max}}}\right)= \frac{(\btheta_{\mathrm{max}}^2 - \btheta_{\mathrm{min}}^2)}{2 \btheta_{\mathrm{max}} \btheta_{\mathrm{min}} \log(\btheta_{\mathrm{max}} / \btheta_{\mathrm{min}})}
    \end{align*}
    \paragraph{\textbf{Case 4:}} $c(\btheta) = \alpha \btheta^2$, $g(z) = z^2$
    \begin{align*}
        \text{CG} &= \left(\frac{\alpha}{(\btheta_{\mathrm{max}} - \btheta_{\mathrm{min}})} \int_{\Theta} \btheta^2 d\btheta\right)  \left(\frac{1}{\alpha^2 (\btheta_{\mathrm{max}} - \btheta_{\mathrm{min}})} \int_{\Theta} \frac{1}{\btheta^4} d\btheta \right) \left(\frac{1}{\frac{1}{\alpha (\btheta_{\mathrm{max}} - \btheta_{\mathrm{min}})} \int_{\Theta} \frac{1}{\btheta^2}d\btheta} \right)\\  
        &= \left(\frac{\left[\btheta^3\right]_{\btheta_{\mathrm{min}}}^{\btheta_{\mathrm{max}}}}{3(\btheta_{\mathrm{max}} - \btheta_{\mathrm{min}})}\right)  \left(-\frac{1}{3} \left[\frac{1}{\btheta^3} \right]_{\btheta_{\mathrm{min}}}^{\btheta_{\mathrm{max}}} \right) \left( \frac{1}{\left[-\frac{1}{\btheta} \right]_{\btheta_{\mathrm{min}}}^{\btheta_{\mathrm{max}}}} \right)= \frac{(\btheta_{\mathrm{max}}^3 - \btheta_{\mathrm{min}}^3)^2}{9 (\btheta_{\mathrm{max}} - \btheta_{\mathrm{min}})^2 \btheta_{\mathrm{max}}^2 \btheta_{\mathrm{min}}^2}
    \end{align*}

\section{Additional experimental details and results}
\label{app:implementation}

\subsection{Implementation details}
\label{sec:implementation_details}
We used the \texttt{sbi} package \citep{Tejero-Cantero2020} (\url{https://sbi-dev.github.io/sbi/}, Version: 0.22.0, License: Apache 2.0) to implement NPE and Ca-NPE. Specifically, we chose the NPE-C model \citep{Greenberg2019} with Masked Autoregressive Flow (MAF) \citep{papamakarios2017masked} as the inference network. We used the default configuration with $50$ hidden units and $5$ transforms for MAF, and training with a fixed learning rate $5 \times 10^{-4}$.
We implemented NLE using the NLE-A model \citep{Papamakarios2019} with the same network configuration as in NPE. For posterior estimation, we used MCMC and chose slice sampling \citep{neal2003slice} as the MCMC method.
For the ABC implementation in the Gamma experiment, we use rejection-ABC algorithm with a tolerance threshold of $\epsilon = 0.05$ and $n=50,000$, in contrast with $n=5000$ for NPE, as rejection-ABC requires much more samples to achieve a reasonable level of posterior approximation. We compute the Euclidean distance between the sample mean and standard deviation of the observed and simulated data, computed from $m=500$ iid samples.
We used a Dell Precision 7550 laptop with Intel i7-10750H processor for running the Gamma experiments and generating the data for the radio propagation model. All the epidemiology experiments are carried out on a CPU cluster consisting of Xeon E5 2680 processors.

\subsection{Maximum mean discrepancy hyperparameters}
\label{sec:mmd_lenghtscale}

We use a Gaussian kernel for implementing MMD, and select the lengthscale parameter using the median heuristic \citep{Garreau2017} computed on the observed dataset (i.e. we take it proportional to the median distance between observations). This is one of the most widely used parameter selection methods for MMD, and we provide the hyperparameter values in \Cref{tab:lengthscale_gamma} and \Cref{tab:lengthscale_SIR}. The amplitude parameter of the kernel is set to 1 as it just acts a multiplicative constant and therefore does not matter so long as it is selected to be the same for all methods.

\begin{multicols}{2}

\begin{table}[H]
    \centering
    \caption{Lengthscales used for the Gamma simulator.}
    \begin{tabular}{lc}
    \toprule
    $\theta_{\mathrm{true}}$ & Lengthscale \\ \midrule
     $ 250$  & 0.48  \\
     $500$  & 0.68 \\
     $750$  & 0.81 \\ \bottomrule
    \end{tabular}
    \label{tab:lengthscale_gamma}
\end{table}

\begin{table}[H]
    \centering
    \caption{Lengthscales used the SIR simulators.}
    \begin{tabular}{lc}
    \toprule
    SIR model & Lengthscale \\ \midrule
     Homogeneous  &  0.10 \\
     Temporal  & 0.19 \\
     Bernoulli  & 0.32 \\ \bottomrule
    \end{tabular}
    \label{tab:lengthscale_SIR}
\end{table}
\end{multicols}

\begin{table}[H]
\centering
\caption{Values of $g_{\mathrm{min}}$ and $g_{\mathrm{max}}$ for the Gamma simulator.}
\scalebox{1}{
\begin{tabular}{@{}rcc@{}}
\toprule
\multicolumn{1}{l}{} & $g_{\mathrm{min}}$ & $g_{\mathrm{max}}$ \\ \midrule
$g(z) = z$           & $10^{-3}$                   &  $2.8 \times 10^{-3}$                  \\
$g(z) = z^2$         & $10^{-6}$                   &  $7.8 \times 10^{-6}$                  \\
$g(z) = z^3$         & $10^{-9}$                   &  $2.2 \times 10^{-8}$                  \\ \bottomrule
\end{tabular}
}
\label{tab:gmin_gamma}
\end{table}

\subsection{Additional results for the Gamma experiment}
\label{app:gamma_additional}

\begin{figure}[H]
    \centering
    \subfigure[]{\includegraphics[width=0.2\textwidth]{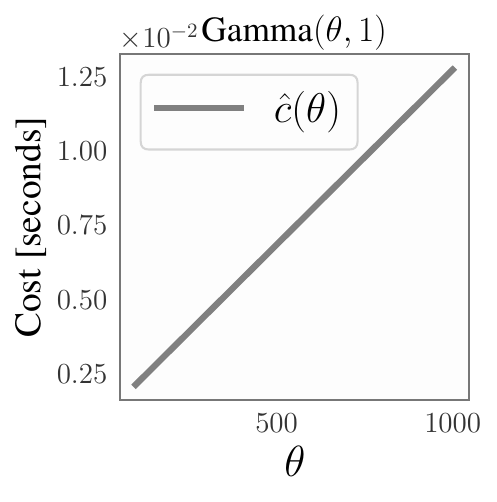}}
    \subfigure[]{\includegraphics[width=0.2\linewidth]{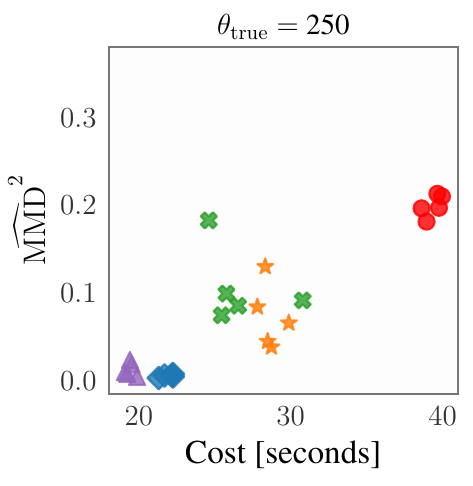}}
    \subfigure[]{\includegraphics[trim={0 0 0 0}, clip, width=0.18\linewidth]{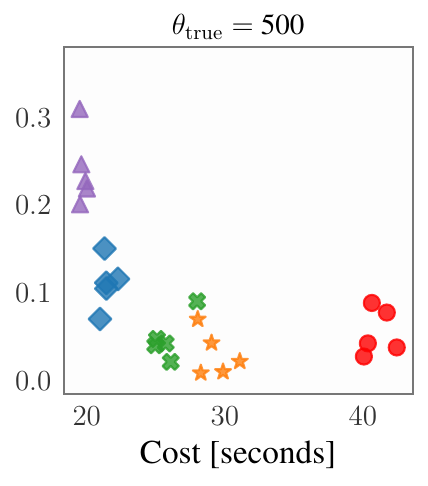}}
    \subfigure[]{\includegraphics[trim={0 0 0 0}, clip, width=0.18\linewidth]{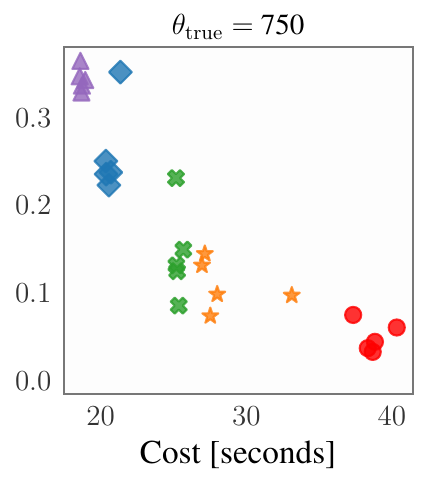}}
    \includegraphics[trim={40 260 30 0}, clip, width=0.17\linewidth]{figures/plot_gamma_legend.pdf}
    \caption{(a) Estimated cost of the Gamma simulator. (b)-(d) MMD between the NPE posteriors and the true posterior for different values of $\theta_{\mathrm{true}}$ over five independent runs with $n=5000$. Sample mean and standard deviation of $m$ points are taken as data.}
    \label{fig:gamma_experiment_npe}
\end{figure}

Here we present the additional results related to the Gamma experiment. We estimate the cost of sampling $m=500$ iid points from the Gamma simulator, averaged over 50 runs, using a linear model in \Cref{fig:gamma_experiment_npe}(a). The parameter $\theta$ is varied in the range $[100,1000]$. The $g_{\mathrm{min}}$ and $g_{\mathrm{max}}$ values for different choices of $g$ is in \Cref{tab:gmin_gamma}.

\begin{figure}[H]
    \centering
    \includegraphics[width=0.18\linewidth]{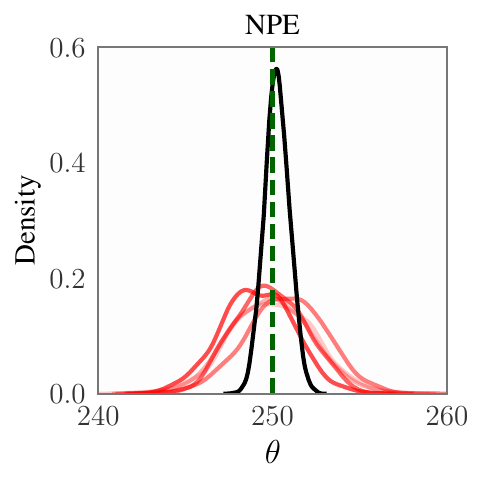}
    \includegraphics[width=0.18\linewidth]{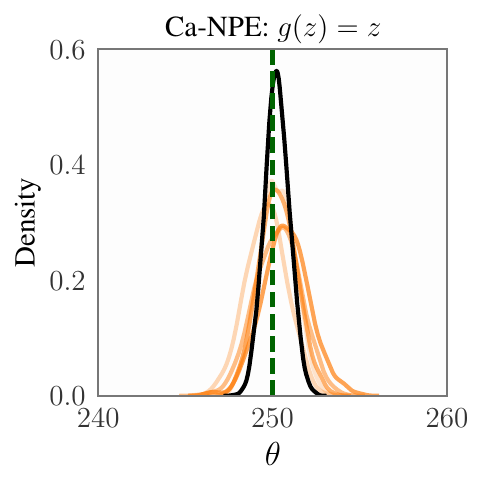}
    \includegraphics[width=0.18\linewidth]{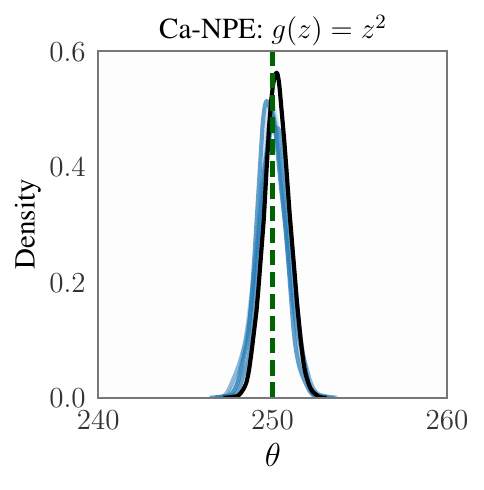}
    \includegraphics[width=0.18\linewidth]{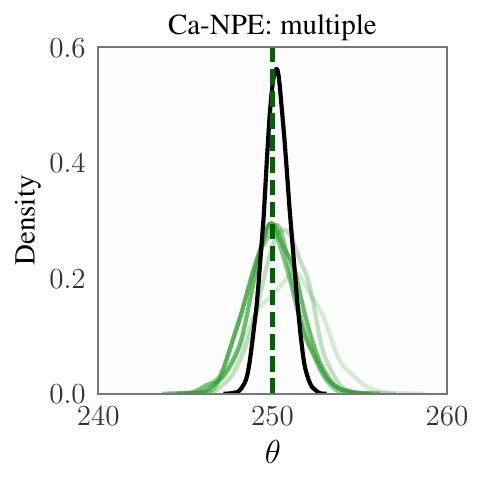}\\
    \includegraphics[width=0.18\linewidth]{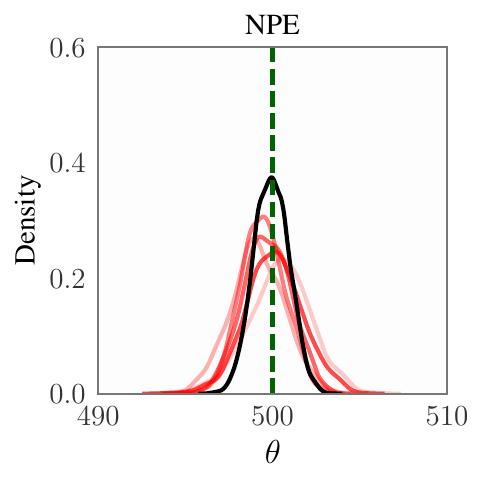}
    \includegraphics[width=0.18\linewidth]{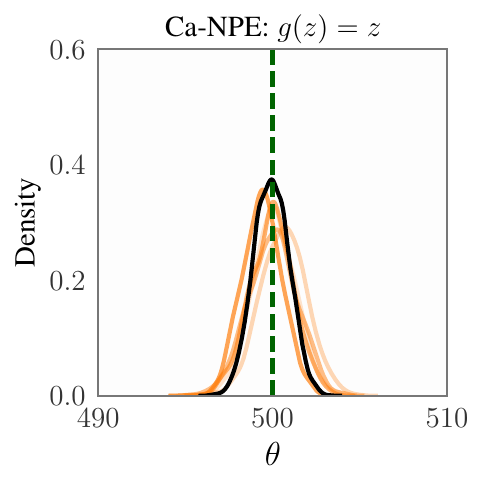}
    \includegraphics[width=0.18\linewidth]{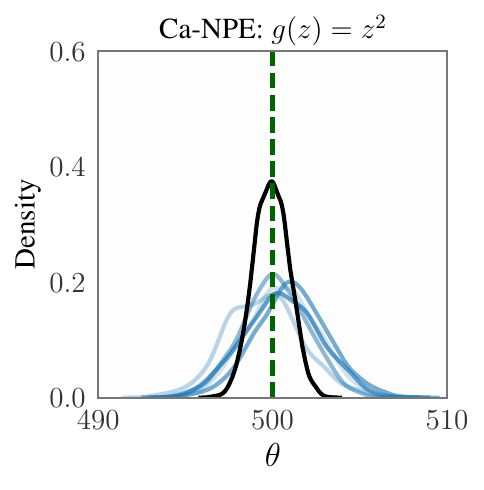}
    \includegraphics[width=0.18\linewidth]{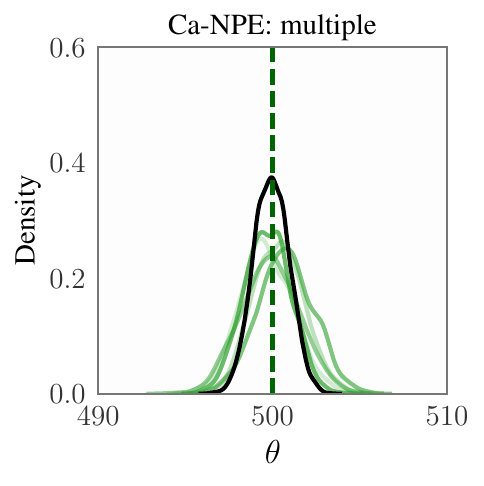}\\
    \includegraphics[width=0.18\linewidth]{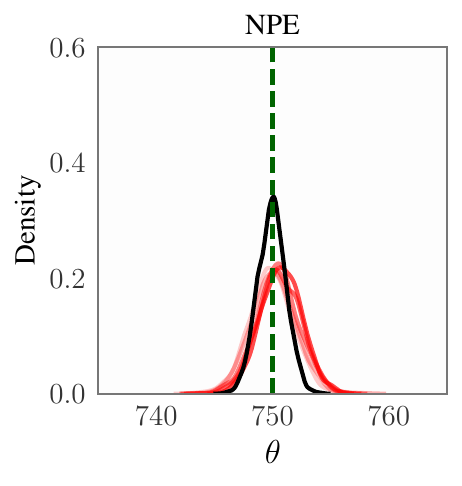}
    \includegraphics[width=0.18\linewidth]{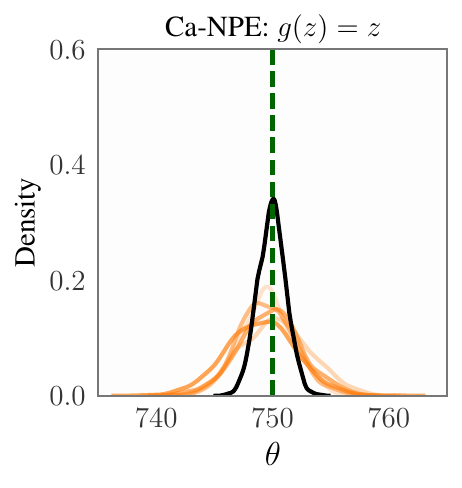}
    \includegraphics[width=0.18\linewidth]{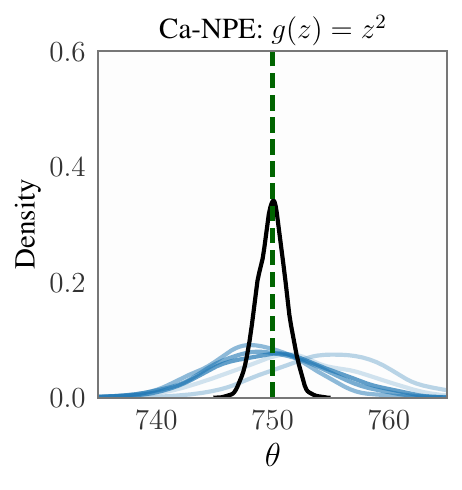}
    \includegraphics[width=0.18\linewidth]{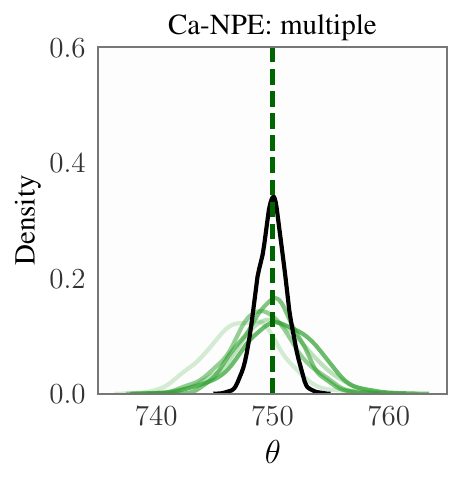}
    \caption{Kernel density estimates of NPE posteriors for the Gamma simulator using $\btheta_{\mathrm{true}} = 250$ (top row), $\btheta_{\mathrm{true}} = 500$ (middle row), and $\btheta_{\mathrm{true}} = 750$ (bottom row) with $n=5000$ samples. The true posterior is in black.} 
    \label{fig:gamma_posterior_npe}
\end{figure}

\begin{figure}[H]
    \centering
    \includegraphics[width=0.18\linewidth]{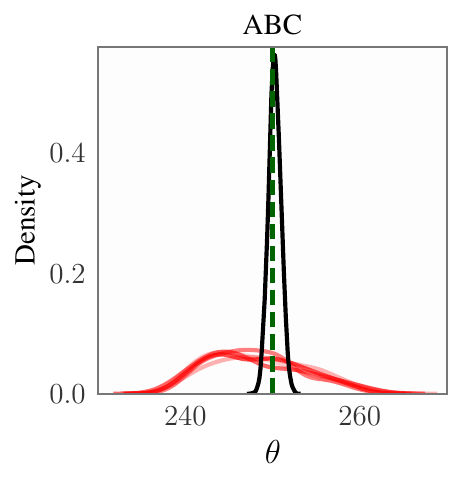}
    \includegraphics[width=0.18\linewidth]{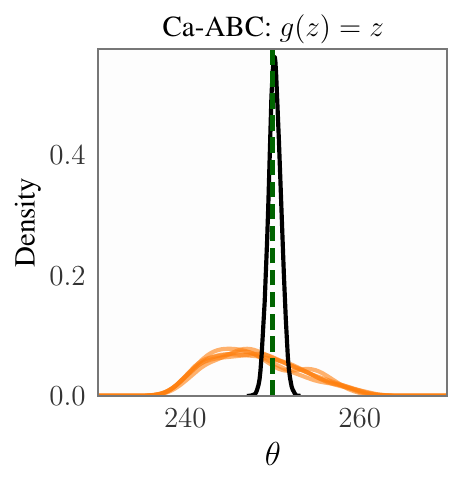}
    \includegraphics[width=0.18\linewidth]{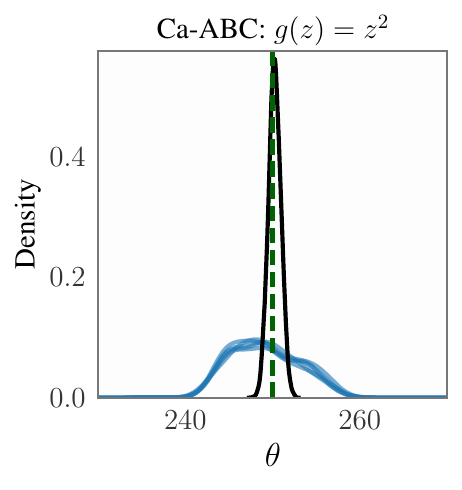}
    \includegraphics[width=0.18\linewidth]{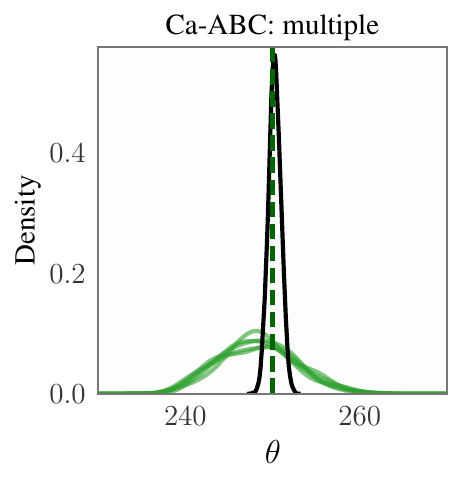}\\
    \includegraphics[width=0.18\linewidth]{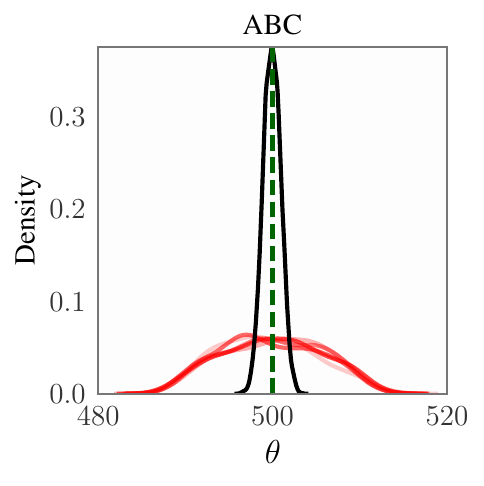}
    \includegraphics[width=0.18\linewidth]{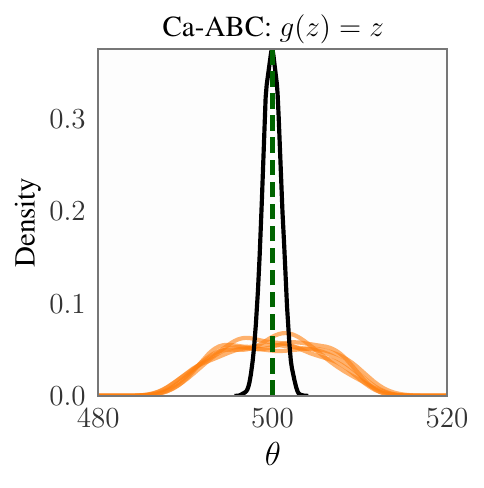}
    \includegraphics[width=0.18\linewidth]{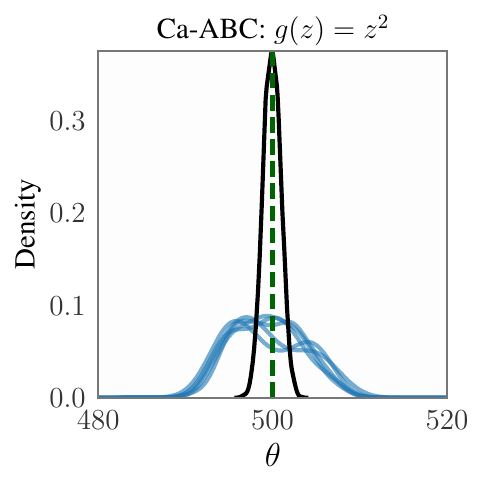}
    \includegraphics[width=0.18\linewidth]{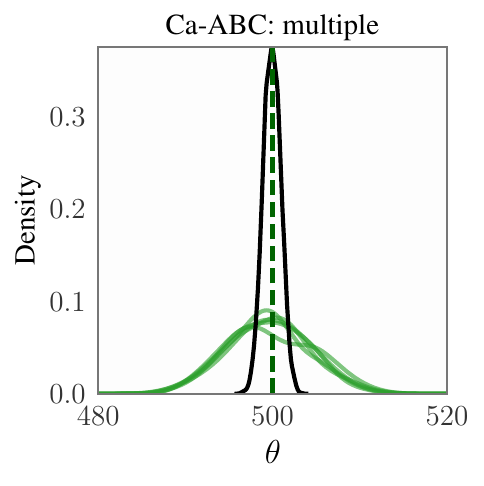}\\
    \includegraphics[width=0.18\linewidth]{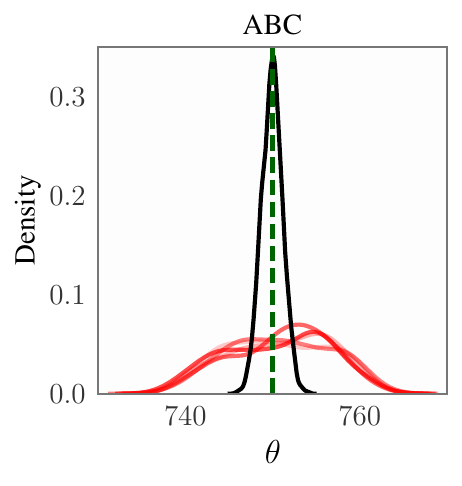}
    \includegraphics[width=0.18\linewidth]{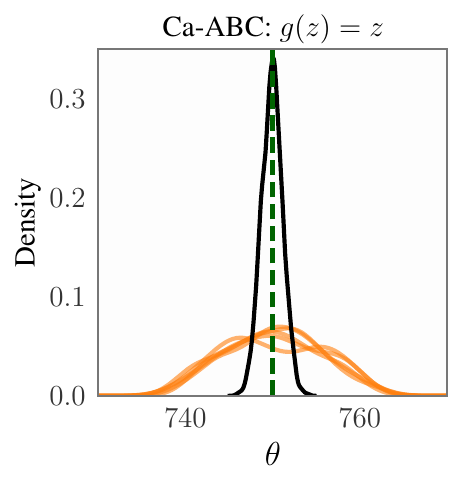}
    \includegraphics[width=0.18\linewidth]{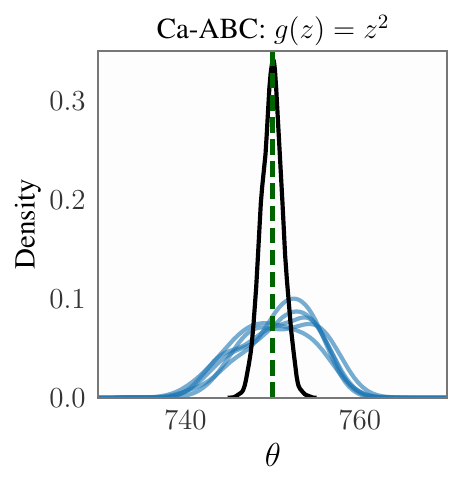}
    \includegraphics[width=0.18\linewidth]{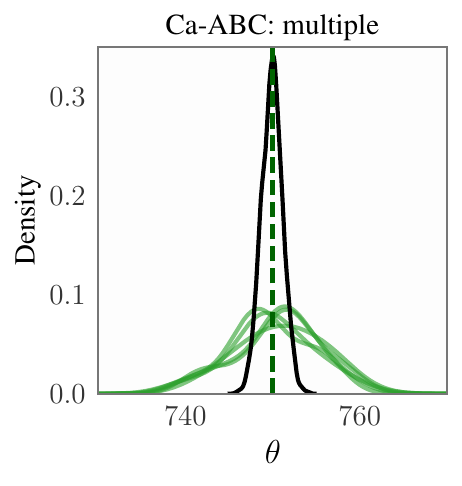}
    \caption{Kernel density estimates of ABC posteriors for the Gamma simulator using $\btheta_{\mathrm{true}} = 250$ (top row), $\btheta_{\mathrm{true}} = 500$ (middle row), and $\btheta_{\mathrm{true}} = 750$ (bottom row) with $n=5000$ samples. The true posterior is in black.} 
    \label{fig:gamma_posterior_ca_abc}
\end{figure}

In \Cref{fig:gamma_experiment_npe}(b)-(d), we plot the MMD between the NPE posteriors and the true posterior for the Gamma simulator, analogous to the ABC plots shown in \Cref{fig:gamma_experiment}. Note that we use $n=5000$ points in the training data for NPE, unlike the $n=50,000$ data points we simulate for ABC, as NPE achieves superior performance than ABC with much fewer samples. Similar to the ABC results, the multiple importance sampling approach achieves similar posterior accuracy as using the prior, whilst reducing  total simulation cost by roughly 25\%. When the true value lies in the low-cost region ($\theta_{\mathrm{true}} = 250$), the NPE posterior approximation improves with increasing penalty exponent $k$, as most of the simulated samples in the training data are from that region. Contrarily, the MMD for NPE increases with $k$ when $\theta_{\mathrm{true}}$ is in the high-cost regions, as the NPE is trained with much fewer samples around $\theta_{\mathrm{true}}$. The approximate posterior distributions obtained using NPE and ABC are shown in \Cref{fig:gamma_posterior_npe} and \Cref{fig:gamma_posterior_ca_abc}, respectively.
%


\subsection{Details and additional results for epidemiology experiments} \label{app:sir_description}

 We considered three models: the homogeneous SIR, the temporal SIR and a Bernoulli SIR. The homogeneous SIR is a simplified model with a single infection rate parameter, where simulation time increases as the rate rises. The temporal SIR model, which accounts for the progression of an epidemic outbreak, has two parameters controlling the infection and recovery rates, both of which impact simulation cost as shown in \Cref{fig:intro}. The Bernoulli SIR is a more complex version based on the temporal SIR, incorporating an additional parameter to control the probability of transmission between individuals. All three parameters of the Bernoulli SIR can influence the simulation cost.

\begin{figure}[H]
    \centering
    \subfigure[]{\includegraphics[width=0.21\linewidth]{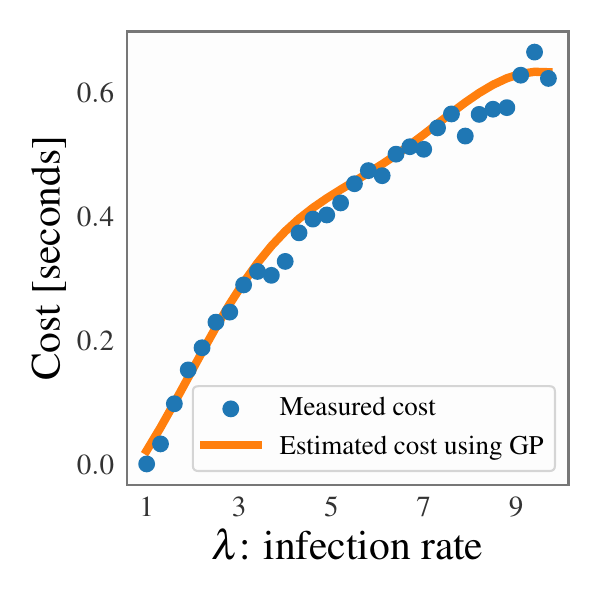}}
    \subfigure[]{\includegraphics[width=0.77\linewidth]{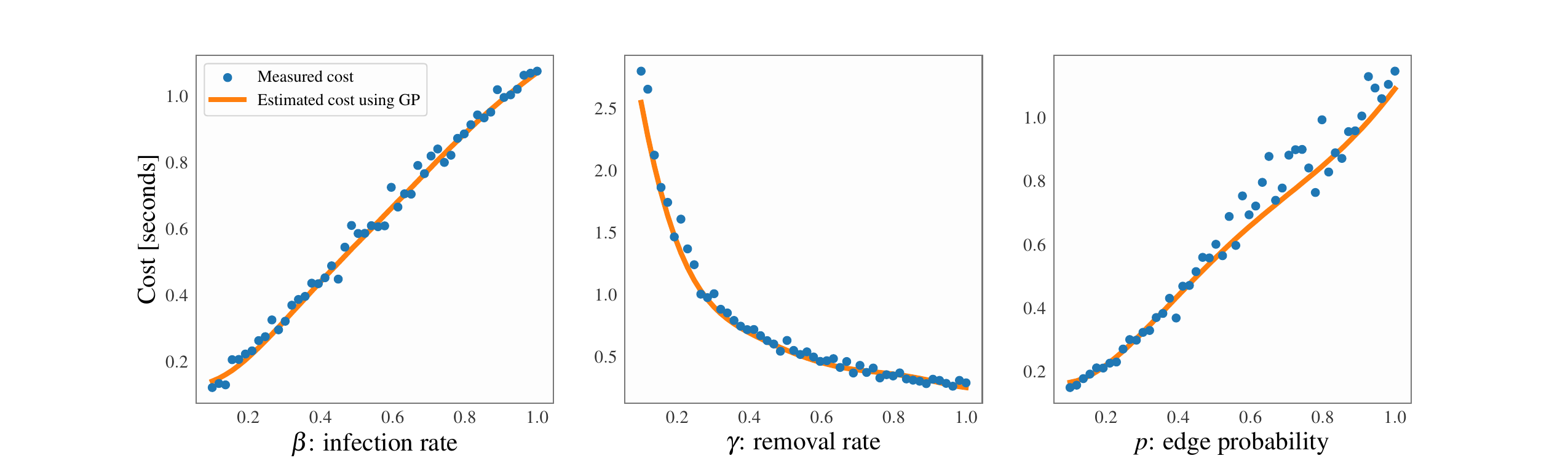}}
    \caption{Costs of sampling one data point from (a) the Homogeneous SIR model, and (b) the Bernoulli SIR model, calculated as the average over 20 simulation runs. The corresponding GP fit for estimating the cost function is shown in orange. The GP is able to estimate the cost function well with limited amount of samples.} 
    \label{fig:cost_bern}
\end{figure}

\textbf{Homogeneous SIR} assumes a closed population where each individual has an equal probability of contacting any other individual. Initially, one individual is infected, and the epidemic spreads through infectious contacts until no infectious individuals remain. The detailed algorithm for this model is described in Algorithm~\ref{alg:homo_sir}. For simulations, the population size $N$ is set to 10,000, and $k=1$. We choose a uniform prior for the infection rate $\theta_1 \sim \mathcal{U}(1, 10)$, and we use $\theta_{\text{true}} = [5]^\top$ to generate the observed data. The estimated simulation costs w.r.t. parameters are shown in Figure~\ref{fig:cost_bern}(a).

\begin{algorithm}[t] 
\caption{Simulation of Homogeneous SIR model}
\label{alg:homo_sir}
\begin{algorithmic}
\item \textbf{Input:} population size $N$, infection rate $\theta_1$, dispersion parameter $k$

\State Initialize $i=1$, $s=N-1$
\While{$i > 0$}
    \State Sample $I \sim \text{Gamma}(k, k)$
    \State Sample $Z \sim \text{Poisson}(\theta_1 \cdot I)$ \Comment{Sample infectious period}
    \For{$1, ..., Z$}
        \State Sample $u \sim \text{Uniform}(0, 1)$
        \If{$u < \frac{s}{N}$}
            \State $s \leftarrow s - 1$
            \State $i \leftarrow i + 1$
        \EndIf
    \EndFor
    \State $i \leftarrow i - 1$ \Comment{individual recovers or dies}
\EndWhile

\item \textbf{Output:} Final epidemic size $N-s$.
\end{algorithmic}
\end{algorithm}

\textbf{Temporal SIR} extends the basic SIR model by incorporating temporal dynamics, where the infection and removal events are tracked over time. This allows for the analysis of the epidemic's progression. The detailed algorithm for this model is described in Algorithm~\ref{alg:temp_sir}. We discretise the outbreak lasted time into a number of bins and calculate the number of removals in each bin as the summary statistics. For simulations, the population size $N$ is set to 1000, and the number of bins is set to $n_b=10$. We choose uniform priors for the infection rate $\theta_1 \sim \mathcal{U}(0.1, 1.0)$ and the removal rate $\theta_2 \sim \mathcal{U}(0.1, 1.0)$, and set $\theta_{\text{true}} = [0.5, 0.5]^\top$ to generate the observed data.

\begin{algorithm}[t] 
\caption{Simulation of Temporal SIR model}
\label{alg:temp_sir}
\begin{algorithmic}
\item \textbf{Input:} population size $N$, number of bins $n_b$, infection rate $\theta_1$, removal rate $\theta_2$

\State Initialize $i=1$, $s=N-1$, $t=0$ 
\State Initialize lists $\mathtt{times} = [t]$, $\mathtt{types} = [1]$ \Comment{1 for infection, 2 for removal}
\While{$i > 0$}
    \State $\tau \leftarrow \text{Exp}(\frac{\theta_1}{N} i s + \theta_2 I)$
    \State Simulate $u \sim \text{Uniform}(0, 1)$
    \If{$u < \frac{\theta_1 s}{\theta_1 s + N \theta_2}$}
        \State $i \leftarrow i + 1$
        \State $s \leftarrow s - 1$
        \State Append $1$ to $\mathtt{types}$
    \Else
        \State $i \leftarrow i - 1$
        \State Append $2$ to $\mathtt{types}$
    \EndIf
    \State $t \leftarrow t + \tau$
    \State Append $t$ to $\mathtt{times}$
\EndWhile

\State $T \leftarrow \text{last element of } \mathtt{times}$
\State Identify the times at which removals occurred from $\mathtt{times}$ based on $\mathtt{types}$
\State Discretise the interval $[0, T]$ into $n_b$ bins and count the number of removals in each bin

\item \textbf{Output:} Final epidemic size $N-s$, duration of the epidemic $T$, removals per bin
\end{algorithmic}
\end{algorithm}

\textbf{Bernoulli SIR} simulates the spread of an epidemic on a network represented by a Bernoulli random graph. Each individual in the network can infect their neighbours with a probability determined by $\theta_3$. The model tracks infection and removal events until the epidemic ends. The detailed algorithm for this model is described in Algorithm~\ref{alg:bern_sir}. For simulations, the population size $N$ is set to 1000, and the number of bins is set to $n_b=10$. We choose uniform priors for the infection rate $\theta_1 \sim \mathcal{U}(0.1, 1.0)$ and the removal rate $\theta_2 \sim \mathcal{U}(0.1, 1.0)$, and the edge probability $\theta_3 \sim \mathcal{U}(0.1, 1.0)$. We use $\theta_{\text{true}} = [0.5, 0.5, 0.5]^\top$ to generate the observed data. The estimated simulation costs w.r.t. parameters are shown in Figure~\ref{fig:cost_bern}(b).

\begin{algorithm}[t] 
\caption{Simulation of Bernoulli SIR model}
\label{alg:bern_sir}
\begin{algorithmic}
\item \textbf{Input:} population size $N$, number of bins $n_b$, infection rate $\theta_1$, removal rate $\theta_2$, edge probability $\theta_3$

\State Initialize $t=0$, $r=0$, set individual 1 infectious and all other individuals susceptibles
\State Initialize list $\mathtt{times} = [t]$
\State Let $\mathcal{I}$ and $\mathcal{S}$ denote the sets of infections and susceptible, respectively.
\State Generate connectivity matrix $G$, where for $i < j, G_{ij}=G_{ji} \sim \text{Bern}(\theta_3)$
\While{$|\mathcal{I}| > 0$}
    \State Simulate $\tau \sim \text{Exp}(\theta_1 \sum_{i\in\mathcal{I}}\sum_{j\in\mathcal{S}}G_{ij}+\theta_2|\mathcal{I}|)$
    \State $t \leftarrow t + \tau$
    \State Simulate $u \sim \text{Uniform}(0, 1)$
    \If{$u < \frac{\theta_1 \sum_{i\in\mathcal{I}}\sum_{j\in\mathcal{S}}G_{ij}}{\theta_1 \sum_{i\in\mathcal{I}}\sum_{j\in\mathcal{S}}G_{ij}+\theta_2|\mathcal{I}|}$}
        \State Sample $J$ from $\mathcal{S}$ with $P(J=j)=\sum_{i\in\mathcal{I}}G_{ij}/\sum_{i\in\mathcal{I}}\sum_{k\in\mathcal{S}}G_{ik}$
        \State $\mathcal{S} \leftarrow \mathcal{S}/\{J\}$, $\mathcal{I} \leftarrow \mathcal{I} \cup \{J\}$
    \Else
        \State Sample $K$ uniformly from $\mathcal{I}$
        \State Remove individual $K$; $\mathcal{I} \leftarrow \mathcal{I} / \{K\}$
        \State Append $t$ to $\mathtt{times}$
        \State $r \leftarrow r+1$
    \EndIf
\EndWhile

\State $T \leftarrow \text{last element of } \mathtt{times}$
\State Discretise the interval $[0, T]$ into $n_b$ bins and count the number of removals in each bin

\item \textbf{Output:} Final epidemic size $r$, duration of the epidemic $T$, removals per bin
\end{algorithmic}
\end{algorithm}

\paragraph{Cost function.}
For epidemiology models, since $c(\theta)$ is not known in advance, we randomly sampled a small dataset $\{(\theta_i, y_i) \}_{i=1}^m$ with $m=200$ parameter and computational time pairs, and estimated it by fitting a Gaussian process (GP) model, see \Cref{fig:cost_bern} for the plots. We used \texttt{GPyTorch} \citep{gardner2018gpytorch} (\url{https://github.com/cornellius-gp/gpytorch}) to implement GP models.

\paragraph{Components for multiple importance sampling.} We use the CG$\times$ESS metric to select the components for the recommended multiple importance sampling scheme, similar to \Cref{table:ess_cost}. Taking the penalty function to be of the form $g(z) = z^k$, we compute CG$\times$ESS for $k=\{0.25, 0.5, 0.75, 1, 1.5, 2\}$, see \Cref{table:ess_cost_sir} for the values. In order to have the same components across all the epidemiology models, we select $k=\{0.5, 1, 2\}$ as the three components apart from the prior. This gives us the right balance between efficiency and cost. The corresponding values for $g_{\mathrm{min}}$ and $g_{\mathrm{max}}$ are reported in \Cref{tab:gmin_sir}. 

\begin{multicols}{2}
\begin{table}[H]
\centering
\caption{CG$\times$ESS for the epidemiology models.}
\scalebox{0.7}{
\begin{tabular}{@{}rclccccccc@{}}
\toprule
\multicolumn{1}{l}{} & SBI   &  & \multicolumn{6}{c}{Ca-SBI}                              \\ \cmidrule(lr){2-2} \cmidrule(l){4-9} 
$g(z)$               & -     &  & $z^{0.25}$ & $z^{0.5}$ & $z^{0.75}$& $z^{1.0}$ & $z^{1.5}$ & $z^2$ \\ \midrule
Homogen.             & 1.0 &  & 1.04      & 1.06       & 1.04         & 0.99  & 0.81 & 0.53 \\
Temporal             & 1.0 &  & 0.93      & 0.87       & 0.73         & 0.56  & 0.26 & 0.14 \\ 
Bernoulli            & 1.0 &  & 1.09      & 1.07       & 0.95         & 0.84  & 0.50 & 0.23 \\ \bottomrule
\end{tabular}
}
\label{table:ess_cost_sir}
\end{table} 

\begin{table}[H]
\centering
\caption{$g_{\mathrm{min}}$ and $g_{\mathrm{max}}$ values for epidemiology models.}
\scalebox{0.7}{
\begin{tabular}{@{}lcclcclcc@{}}
\toprule
          & \multicolumn{2}{c}{$g(z) = z^{0.5}$}    &  & \multicolumn{2}{c}{$g(z) = z$}          &  & \multicolumn{2}{c}{$g(z) = z^2$}        \\
          & $g_{\mathrm{min}}$ & $g_{\mathrm{max}}$ &  & $g_{\mathrm{min}}$ & $g_{\mathrm{max}}$ &  & $g_{\mathrm{min}}$ & $g_{\mathrm{max}}$ \\ \cmidrule(lr){2-3} \cmidrule(lr){5-6} \cmidrule(l){8-9} 
Homogen.  &   0.1498                 &   0.7929                 & & 0.0225 &    0.6288               & &  0.0005                  & 0.3953                                        \\
Temporal  &  0.0930                  &   0.3250                 &  &  0.0087                  &   0.1056                 &  &   0.0001                 &  0.0112                  \\
Bernoulli &   0.1197                 &  0.9190                  &  &   0.0143                 &   0.8447                 &  &  0.0002                  &   0.7134                 \\ \bottomrule
\end{tabular}
}
\label{tab:gmin_sir}
\end{table}
\end{multicols}

\paragraph{Additional performance metrics.} Apart from the MMD, we also report additional performance metrics such as the classifier 2-sample test (C2ST) \citep{friedman2003multivariate, lopez-paz2017revisiting} and the marginal two-sample Kolmogorov-Smirnov (KS) \citep{hodges1958significance} test scores for the SIR experiments. C2ST involves training a classifier to distinguish between two distributions, which in our case is the reference posterior and the inferred posterior. The test outputs the classification accuracy, shown in \Cref{tab:C2ST}, which at best is 0.5 in case the two distributions are the same. The two-sample KS test statistic, reported in is \Cref{tab:ks_test}, quantifies the distance between the empirical cumulative distribution functions of two distributions. Our conclusions remain the same. 

\begin{table}[H]
\centering
\caption{C2ST scores ($\downarrow$) for NPE and Ca-NPE on the three SIR models. The mean and \textcolor{gray}{standard deviation} from $50$ runs are reported.}
\scalebox{0.9}{
\begin{tabular}{rccccc}
\hline

                & NPE                             & Ca-NPE $(g(z) = z^{0.5})$ & Ca-NPE $(g(z) = z)$ & Ca-NPE $(g(z) = z^{2})$ & Ca-NPE (multiple) \\ \hline
Homogen. & $ 0.56 \textcolor{gray}{(0.04)}$ & $0.55 \textcolor{gray}{(0.02)}$                                                                  & $0.55 \textcolor{gray}{(0.03)}$                                                          & $0.77 \textcolor{gray}{(0.06)}$                                    &   $ 0.59 \textcolor{gray}{(0.04)}$   \\
Temporal    & $0.60 \textcolor{gray}{(0.03)}$ & $0.66 \textcolor{gray}{(0.05)}$                                   & $0.70 \textcolor{gray}{(0.03)}$ & $0.75 \textcolor{gray}{(0.04)}$                                                                          &     $0.65 \textcolor{gray}{(0.03)}$                                                            \\
Bernoulli   & $0.93 \textcolor{gray}{(0.01)}$                                & $0.93 \textcolor{gray}{(0.01)}$                                                                  & $0.93 \textcolor{gray}{(0.01)}$                                                          & $0.92 \textcolor{gray}{(0.02)}$                                                              &  $ 0.92 \textcolor{gray}{(0.02)}$                                                                             \\ \hline
\end{tabular}
}
\label{tab:C2ST}
\end{table}

\begin{table}[H]
\centering
\caption{Two-sample KS test scores ($\downarrow$) for NPE and Ca-NPE on the three SIR models. The mean and \textcolor{gray}{standard deviation} from $50$ runs are reported. }
\scalebox{0.81}{
\begin{tabular}{rccccc}
\hline
                
                & NPE                             & Ca-NPE $(g(z) = z^{0.5})$ & Ca-NPE $(g(z) = z)$ & Ca-NPE $(g(z) = z^{2})$ & Ca-NPE (multiple) \\ \hline
Homogen. & $ 0.12 \textcolor{gray}{(0.05)}$ & $0.11 \textcolor{gray}{(0.04)}$                                                                  & $0.11 \textcolor{gray}{(0.05)}$                                                          & $0.49 \textcolor{gray}{(0.14)}$                                    &   $ 0.15 \textcolor{gray}{(0.06)}$     \\
Temporal    & \begin{tabular}[c]{@{}c@{}}{[}$0.12 \textcolor{gray}{(0.06)}$,\\  $0.12 \textcolor{gray}{(0.06)}${]}\end{tabular} & \begin{tabular}[c]{@{}c@{}}{[}$0.11 \textcolor{gray}{(0.05)}$, \\ $0.13 \textcolor{gray}{(0.07)}${]}\end{tabular}                                  & \begin{tabular}[c]{@{}c@{}}{[}$0.13 \textcolor{gray}{(0.05)}$, \\ $0.13 \textcolor{gray}{(0.07)}${]}\end{tabular} & \begin{tabular}[c]{@{}c@{}}{[}$0.13 \textcolor{gray}{(0.07)}$, \\ $0.19 \textcolor{gray}{(0.07)}${]}\end{tabular}                                                                         &     \begin{tabular}[c]{@{}c@{}}{[}$0.12 \textcolor{gray}{(0.04)}$, \\ $0.13 \textcolor{gray}{(0.06)}${]}\end{tabular}                                                                          \\
Bernoulli   & \begin{tabular}[c]{@{}c@{}}{[}$0.12 \textcolor{gray}{(0.03)}$, \\ $0.06 \textcolor{gray}{(0.03)}$, \\ $0.08 \textcolor{gray}{(0.03)}${]}\end{tabular}                             & \begin{tabular}[c]{@{}c@{}}{[}$0.13 \textcolor{gray}{(0.03)}$, \\ $0.05 \textcolor{gray}{(0.02)}$, \\ $0.09 \textcolor{gray}{(0.02)}${]}\end{tabular}                                                                & \begin{tabular}[c]{@{}c@{}}{[}$0.14 \textcolor{gray}{(0.03)}$, \\ $0.07 \textcolor{gray}{(0.03)}$, \\ $0.09 \textcolor{gray}{(0.03)}${]}\end{tabular}                                                     & \begin{tabular}[c]{@{}c@{}}{[}$0.15 \textcolor{gray}{(0.03)}$, \\ $0.13 \textcolor{gray}{(0.03)}$, \\ $0.12 \textcolor{gray}{(0.04)}${]}\end{tabular}                                                        &   \begin{tabular}[c]{@{}c@{}}{[}$0.12 \textcolor{gray}{(0.03)}$, \\ $0.05 \textcolor{gray}{(0.02)}$, \\ $0.10 \textcolor{gray}{(0.04)}${]}\end{tabular}        \\ \hline
\end{tabular}
}
\label{tab:ks_test}
\end{table}

\subsection{Details of the radio propagation experiment} 
\label{app:radio_propagation}
The radio propagation model used in \Cref{sec:radio_experiment} is a stochastic of how the environment affects the transmission of radio signals between a transmitter and a receiver. By adjusting the parameters of this model, one can simulate different communication environments, such as a small room, a large hall or a corridor, which then helps in testing new wireless communication systems. A pseudocode for simulating from this model is given in \Cref{alg:radio}, see \citet{Huang2023} for a more detailed description.

\begin{figure}[H]
    \centering
    \caption{Cost of sampling $m=50$ iid realisations from the radio propagation model as a function of $\theta_3$.  } 
    \includegraphics[width=0.4\linewidth]{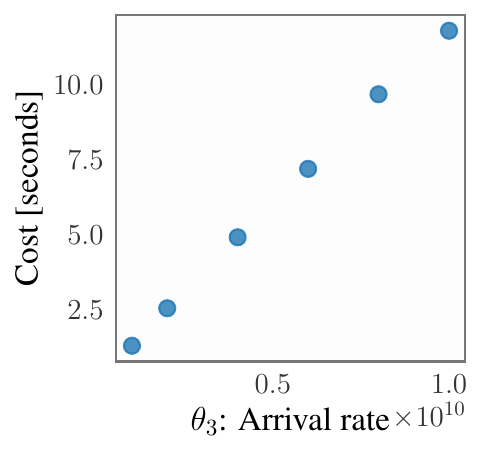}
    \label{fig:cost_turin}
\end{figure}

This model has four parameters, out of which, only $\theta_3$ affects the computational cost, as shown in \Cref{fig:cost_turin}. We use a linear model to estimate the cost function. For quadratic penalty function $g(z) = z^2$, $g_{\mathrm{min}} = 1.7$ and $g_{\mathrm{max}} = 139.2$. Again, we use the CG$\times$ESS metric to select the components for the multiple importance sampling variant of our method, see \Cref{tab:CG_ess_radio} for the values. We see the $g(z) = z$ yields the same CG$\times$ESS as the standard SBI method, which means that we do not lose out on efficiency with this choice of $g$. Hence, we take the cost-aware components to be $\{z, z^2, z^3\}$ in order to get computational advantages from the higher powers.

\begin{figure}[H]
    \centering
    \includegraphics[width=\textwidth]{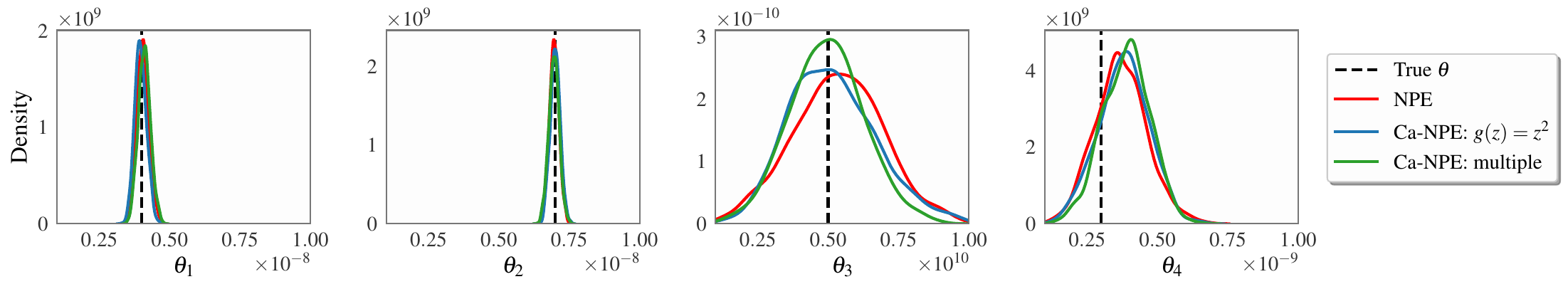}
    \caption{Marginals of the approximate posterior for the radio propagation model using NPE and our Ca-NPE method with $n=10,000$. The Ca-NPE methods perform similar to NPE.}
    \label{fig:turin_npe}
\end{figure}

\Cref{fig:turin_npe} shows the NPE results for the radio propagation model, analogous to the NLE results of \Cref{fig:turin_nle} from the main text. The observations remain the same: the posteriors obtained from NPE and Ca-NPE (for both multiple cost-aware importance sampling and $g(z) = z^2$) are similar, indicating that Ca-NPE reduces cost without sacrificing posterior accuracy.

\begin{multicols}{2}

\begin{table}[H]
\centering
\caption{CG$\times$ESS for the radio propagation model.}
\scalebox{0.9}{
\begin{tabular}{@{}rclcccc@{}}
\toprule
\multicolumn{1}{l}{} & SBI   &  & \multicolumn{4}{c}{Ca-SBI}                              \\ \cmidrule(lr){2-2} \cmidrule(l){4-7} 
$g(z)$               & -     &  & $z^{0.5}$ & $z$ & $z^2$ & $z^3$ \\ \midrule
CG$\times$ESS             & 1.0 &  & 1.06      & 1.0        & 0.60         & 0.22 \\ \bottomrule
\end{tabular}
}
\label{tab:CG_ess_radio}
\end{table} 
\end{multicols}

\begin{algorithm}[H] 
\caption{Simulation of the radio propagation model}
\label{alg:radio}
\begin{algorithmic}
\item \textbf{Input:} Bandwidth $B= 4$ GHz, length of time-domain signal $N_s = 801$, $m=50$ realisations, reverberation gain $\theta_1$, reverberation time $\theta_2$, arrival rate $\theta_3$, noise variance $\theta_4$

\State Set $\Delta f = B/(N_s-1)$ and $t_{\mathrm{max}} = 1/ \Delta f$
\While{$i \leq m $}
    \State Sample $N_{\mathrm{points}} \sim \text{Poisson}(\theta_3 t_{\mathrm{max}})$
    \State Sample $\tau_j \sim \mathcal{U}(0, t_{\mathrm{max}})$, $j=1,\dots, N_{\mathrm{points}}$ and arrange them in ascending order
    \State Sample $\beta_j \sim \mathcal{CN}(0, \sigma^2_P(\tau_j))$, $j=1,\dots, N_{\mathrm{points}}$, where $\sigma^2_P(\tau) = \theta_1 \exp(- \tau / \theta_2) / \theta_3 B$ 
    \State Compute $H_l = \sum_{j=1}^{N_{\mathrm{points}}}\beta_j\exp(-i2\pi l \Delta f \tau_j) $, $l=1, \dots, N_s$ (here, $i$ refers to the imaginary unit)
    \State Sample $W_l \sim \mathcal{CN}(0, \theta_4)$ and compute $Y_l = H_l + W_l$, $l=1, \dots, N_s$
    \State Take inverse Fourier transform of $Y_1, \dots, Y_{N_s}$ to obtain the time-domain signal $y_1, \dots, y_{N_s}$
    \State Compute the temporal moments as $M_{ik} = \int t^k y(t) d t$ for $k = 0,1,2$
\EndWhile
\State Compute the sample mean and sample variance of $\{M_{ik}\}_{i=1}^m$.
\item \textbf{Output:} Data vector $\mathbf{x} \in \mathbb{R}^6$ consisting of three means and three variance estimates of temporal moments.
\end{algorithmic}
\end{algorithm}
%

\end{document}